%% file: example_paper.tex
\documentclass{article}

\usepackage{microtype}
\usepackage{graphicx}
\usepackage{subfigure}
\usepackage{booktabs} 
\usepackage{svg}

\usepackage{hyperref}
\usepackage[most]{tcolorbox}
\usepackage[dvipsnames]{xcolor}
\definecolor{customblue}{rgb}{0.33, 0.5, 0.65}

\usepackage[preprint]{icml2026}

\usepackage{amsmath}
\usepackage{amssymb}
\usepackage{mathtools}
\usepackage{amsthm}

\usepackage[capitalize,noabbrev]{cleveref}

\theoremstyle{plain}

\theoremstyle{definition}

\theoremstyle{remark}

\usepackage[textsize=tiny]{todonotes}
\usepackage{csquotes}
\usepackage{xcolor}

\usepackage{enumitem}
\newcommand{\llamaeightb}{Llama-3.1-8B}
\newcommand{\gemmaoneb}{Gemma-3-1B}
\newcommand{\gemmalarge}{Gemma-27-B}
\newcommand{\qwen}{Qwen-3-32B}

\newlength{\myskip}
\icmltitlerunning{LLM Judges Fail to Reliably Measure Adversarial Robustness}

\begin{document}

\twocolumn[




\icmltitle{
A Coin Flip for Safety: LLM Judges Fail to Reliably Measure \\Adversarial Robustness}


\icmlsetsymbol{equal}{*}

\begin{icmlauthorlist}
\icmlauthor{Leo Schwinn}{tum,mila,equal}
\icmlauthor{Moritz Ladenburger}{tum,equal}
\icmlauthor{Tim Beyer}{tum}
\icmlauthor{Mehrnaz Mofakhami}{mila}
\icmlauthor{Gauthier Gidel}{mila,montreal,cifar}
\icmlauthor{Stephan G\"unnemann}{tum}
\end{icmlauthorlist}

\icmlaffiliation{tum}{Technical University of Munich}
\icmlaffiliation{mila}{Mila}
\icmlaffiliation{montreal}{Universit\'{e} de Montr\'{e}al}
\icmlaffiliation{cifar}{Canada AI CIFAR Chair}

\icmlcorrespondingauthor{Leo Schwinn}{l.schwinn@tum.de}

\icmlkeywords{Machine Learning, ICML, LLM robustness, judging, autograder}

\vskip 0.3in
]



\printAffiliationsAndNotice{\icmlEqualContribution} 

\begin{abstract}
Automated \enquote{LLM-as-a-Judge} frameworks have become the de facto standard for scalable evaluation across natural language processing. For instance, in safety evaluation, these judges are relied upon to evaluate harmfulness in order to benchmark the robustness of safety against adversarial attacks. However, we show that existing validation protocols fail to account for substantial distribution shifts inherent to red-teaming: diverse victim models exhibit distinct generation styles, attacks distort output patterns, and semantic ambiguity varies significantly across jailbreak scenarios. Through a comprehensive audit using 6642 human-verified labels, we reveal that the unpredictable interaction of these shifts often causes judge performance to degrade to near random chance. This stands in stark contrast to the high human agreement reported in prior work. Crucially, we find that many attacks inflate their success rates by exploiting judge insufficiencies rather than eliciting genuinely harmful content. To enable more reliable evaluation, we propose ReliableBench, a benchmark of behaviors that remain more consistently judgeable, and JudgeStressTest, a dataset designed to expose judge failures. Data available at: \url{https://github.com/SchwinnL/LLMJudgeReliability}.
\end{abstract}

\input{chapter/1_introduction}
\input{chapter/2_related_work}

\input{chapter/3_method}

\input{chapter/4_results}
\input{chapter/5_conclusion}


\section*{Impact Statement}

We study the reliability of LLM judges in adversarial safety evaluations. Our research has the potential to improve the reliability of future judges and to yield scientific insights from safety evaluations conducted with novel attacks and defenses. However, our work also has potential limitations and downstream impacts. By relying heavily on the HarmBench dataset, we may inadvertently propagate its inherent biases, such as a focus on single-turn textual safety and specific harm categories, while neglecting other critical aspects like multi-turn agentiness or multimodal safety. Furthermore, our exclusive focus on adversarial robustness might prioritize this metric over other important safety dimensions. Finally, by demonstrating the unreliability of current safety measurement tools, this research impacts the political discourse on AI safety and future strategic priorities, highlighting the urgent need for more robust evaluation standards before deploying autonomous systems in high-stakes environments.

\bibliography{example_paper}
\bibliographystyle{icml2026}

\newpage
\appendix
\onecolumn
\input{chapter/appendix}

\end{document}

%% file: chapter/1_introduction.tex

\section{Introduction}

The widespread deployment of Large Language Models (LLMs) necessitates rigorous safety alignment to prevent the generation of harmful content. As red-teaming efforts scale to identify vulnerabilities, manual human evaluation has become prohibitively expensive, leading to a reliance on "LLM-as-a-Judge" frameworks~\cite{zou2023universal}. Here, LLMs are used as semantic classifiers to assess whether a given generation meets predefined criteria, i.e., whether an output is harmful. LLM judges are typically validated by comparing their agreement with human ratings on static held-out test sets and generally achieve very high human rater agreement~\cite{mazeika2024harmbench, inan2023llama}.

Yet, we argue that current validation protocols of safety judges fundamentally mismatch one of their primary use cases in academia: \textit{adversarial} safety evaluations~\cite{mazeika2024harmbench, souly2024strongreject}. While judges are validated on both benign and harmful model outputs, their initial assessment does not consider relevant distribution shifts that occur in standard adversarial attack~\cite{zou2023universal, ludke2025diffusion} and robustness evaluations~\cite{schwinn2025adversarial}. Namely, we identify three critical distribution shifts that undermine judge reliability in practice. First, \textit{Attack Shift} occurs because adversarial prompts may induce distorted and high-perplexity outputs that differ significantly from the standard harmful responses judges are trained to recognize~\cite{li2025llms}. Second, \textit{Model Shift} arises when a judge validated on the outputs of one model is applied to diverse defenses and models, introducing linguistic shifts that degrade classification performance. Third, \textit{Data Shift}, where judging difficulty varies significantly by semantic category. For example, subtle propaganda may be considerably harder to reliably detect compared to explicit violence.

\begin{figure}
    \centering
    \subfigure{
        \includegraphics[width=1\linewidth]{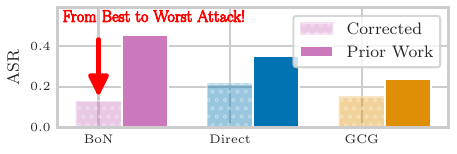}
    }
    \caption{Impact of judge unreliability on reported Attack Success Rates (ASR). Standard automated judges significantly overestimate the success of adversarial attacks compared to human verification.}
    \label{fig:best_to_worst}
    \vspace{-0.7cm}
\end{figure}

We argue that these failure modes severely compromise the validity of safety research. As Figure~\ref{fig:best_to_worst} shows, reported ASR is often severely inflated; we obtain \textit{corrected} ASR by scaling results by the judge's precision, the probability that a judge-positive is a true positive. When judges are affected by unknown distribution shifts, it becomes difficult to discern genuine progress from measurement error. A reported increase in attack success may simply reflect an attack exploiting a judge's False Positive Rate, where positive indicates a successful attack. Conversely, a model's apparent robustness may also stem from response styles that trigger False Negatives. Thus, comparative analyses of attack costs and model robustness may become artifacts of judge unreliability rather than measures of genuine safety. We emphasize the following:
\begin{center}
\emph{LLM judges are substantially less reliable for adversarial safety evaluations than previously assumed, performing on average only slightly better than a random coin-flip.}
\end{center}
In this paper, we conduct an extensive audit of LLM judges in typical adversarial evaluation settings using manual human labeling as a gold standard. We analyze failure modes across varying attacks, victim models, and semantic behavior categories, quantifying how reliable judges are under realistic evaluation settings and how reported attack success rates change when correcting for judge errors. Our contributions can be summarized as follows:

\textbf{Auditing Judge Reliability with Human Labels.} We conduct a rigorous evaluation of LLM-based safety judges using a high-quality dataset of 6,442 human-verified samples, validated through strict labeling protocols. We demonstrate that distribution shifts present in adversarial robustness evaluations, induced by attacks, victim-model differences, and the semantic complexity of judged behaviors, degrade judge performance to near-random chance.

\textbf{Impact on Safety Evaluation Validity.} We quantify how judge failures inflate reported Attack Success Rates (ASR). We show that popular attacks, such as Best-of-N (BoN), disproportionately exploit judge false positives rather than model vulnerabilities. As illustrated in Figure~\ref{fig:best_to_worst}, correcting for these errors can considerably change the perceived effectiveness of attacks.

\textbf{More Reliable Evaluations.} We provide actionable strategies to mitigate judge unreliability. We demonstrate that (1) evaluations become more reliable when collecting multiple judge-positive samples per behavior, (2) correcting ASR for judge precision yields more realistic robustness estimates (as in Figure~\ref{fig:best_to_worst}), and (3) filtering for consistent behaviors improves judge reliability. To support this, we release \textbf{ReliableBench}, a curated subset of \enquote{easy-to-judge} behaviors, and \textbf{JudgeStressTest}, a dataset of hard failure cases to facilitate research into robust judges.

%% file: chapter/2_related_work.tex
\section{Related work}

\textbf{LLM-as-a-Judge for Safety Evaluation.} In most domains, adversarial robustness can be directly assessed due to the simplicity of classification tasks~\cite{madry_towards_2018, altstidl2024scalability}. However, the open-endendess of natural language processing complicates this procedure. Early safety evaluation relied on static datasets and classifiers, such as RealToxicityPrompts~\cite{gehman2020realtoxicityprompts} and SOLID~\cite{rosenthal2021solid}, which provided benchmarks for toxic content detection. As LLMs scaled, the field shifted toward using LLMs themselves as evaluators. This \enquote{LLM-as-a-Judge} paradigm~\cite{zheng2023judging} demonstrated that strong models like GPT-4 can achieve high agreement with human raters on general quality assessments. For safety-specific evaluation, specialized systems emerged: Llama Guard~\cite{inan2023llama} introduced fine-tuned safeguard models, AEGIS~\cite{ghosh2024aegis} proposed ensemble approaches across 13 risk categories, and WildGuard~\cite{han2024wildguard} achieved state-of-the-art open-source performance, reducing jailbreak success rates compared to prior methods. Moreover, standardized benchmarks such as HarmBench~\cite{mazeika2024harmbench} have enabled systematic comparisons of attacks and defenses.

\textbf{Adversarial Vulnerabilities of LLM Judges.}
Recent work has demonstrated that LLM judges are actively vulnerable to adversarial manipulation. \textit{Judge jailbreaking} occurs when prompt injections trick the evaluator into outputting favorable verdicts regardless of content~\cite{maloyan2025adversarial}. \textit{Adversarial confusion} exploits judges through high-perplexity suffixes or obfuscated inputs that confound analysis. ~\citet{li2025llms} evaluated 15 attack techniques in their RobustJudge framework, finding that optimization-based attacks could cause judges to misclassify 100\% of harmful outputs as safe. While defenses such as input retokenization and ensemble voting provide partial mitigation~\cite{li2025llms}, these findings establish that current judges can be systematically defeated by prompt-level attacks.

\textbf{Failure Modes and Distribution Sensitivity.}
Beyond active attacks, judges exhibit systematic failures under distribution shift.~\citet{feuer2024style} demonstrated that judges prioritize stylistic features, such as completeness, politeness, formatting, over factual correctness or safety.~\citet{chen2025safer} quantified this phenomenon, showing that adding an apologetic prefix alone shifted judge preferences by up to 98\%, enabling models to \enquote{hack} safety metrics through tone. Other works found that minor stylistic perturbations can increase false negative rates substantially~\cite{zheng2023judging, eiras2025know}. 

In addition to sensitivity to style, \citet{Liu2025Calibration} identified critical failure modes in judge calibration. Their evaluation across several guard models revealed that judge models tend to make overconfident predictions, and that miscalibration worsens under distribution shift, such as when subjected to jailbreak attacks or responses coming from different models.
Furthermore, recent work exposed that judges face significant challenges in evaluating the semantic validity of harmful outputs and demonstrated that they rely more on malicious linguistic patterns than authentic harmful knowledge~\citep{mei20240not, yan2025confusion}. They showed that a large fraction of alleged jailbreak successes are actually hallucinated or practically useless, harmful instructions, leading evaluators to misrepresent real-world risk. Furthermore,~\citet{panfilov2025strategic} identify strategic dishonesty as a key limitation of judges, where models generate deceptive, harmful-sounding outputs that fool safety classifiers, compounding the reliability issues we observe under adversarial attack shifts.

\textbf{Gaps in Prior Work.}
While existing research has identified individual failure modes in LLM-as-a-judge frameworks, critical gaps remain. First, no prior work systematically investigates the effect of relevant distribution shifts in adversarial safety evaluations (attack, model, and data shift), leaving their joint effects unexplored. Second, existing evaluations lack human-labeled ground truth to quantify which semantic or functional categories pose genuine challenges versus systematic biases. Third, within attack shift, the problem of \textit{judge-hacking}, whether through optimization-based attacks that maximize judge scores or sampling-based strategies that accumulate false positives by generating many candidates, remains unexamined despite posing a realistic threat to evaluation integrity. Thus, it remains unclear whether new attacks achieve genuinely higher success rates or are simply better at exploiting judges, as seen in Figure~\ref{fig:best_to_worst}.

%% file: chapter/3_method.tex
\section{Method}

Here, we describe the setup used in our empirical study.

\subsection{Initial Dataset} 

To systematically investigate how attack, victim model, and data shifts affect judge reliability, we require a diverse set of harmful queries with well-defined semantic categories. We base our labeling efforts on the HarmBench test dataset~\cite{mazeika2024harmbench} for two primary reasons: (1) it is well-established in the adversarial robustness community, facilitating comparability with prior work, and (2) it provides fine-grained annotations of semantic and functional categories, enabling us to analyze category-dependent variations in judge performance (i.e., data shift).
The original HarmBench test set comprises 400 unique harmful queries. However, since our study aims to characterize judge behavior across multiple attacks, victim models, and semantic categories, labeling all 400 queries for every combination would yield prohibitively few samples per experimental condition, potentially compromising reliability. To balance coverage across our experimental conditions while maintaining sufficient sample sizes per condition, we first randomly subsample 100 queries from HarmBench. This subsampling allows us to allocate labeling resources across a broader range of attack–model combinations while retaining enough examples per setting to draw reliable conclusions.

\subsection{Victim Models} 

To investigate model shift, specifically, whether model family and model size influence judge errors, we evaluate across a diverse set of 4 open-weight models spanning different architectures and scales. For smaller models, we include \gemmaoneb~\cite{team2025gemma} and \llamaeightb{}~\cite{grattafiori2024llama}. For the larger model, we employ a \gemmalarge~\cite{team2025gemma} and \qwen~\cite{yang2025qwen3}. This selection enables us to disentangle the effects of model capacity from architectural differences on judge reliability. While newer frontier open-weights models have been released since the inception of this study, we selected a representative set of models that were state-of-the-art when we conducted the human labeling and allow for efficient large-scale evaluation. Our focus lies on identifying fundamental mechanisms of judge failure under distribution shift, which we expect to generalize across model generations. We deliberately exclude proprietary models from our evaluation for two reasons. First, proprietary models are frequently updated or deprecated without notice, undermining long-term reproducibility, a critical concern for benchmark studies intended to support future research~\cite{schwinn2023adversarial, beyer2025llm}. Second, we observed that submitting harmful queries for judgment frequently triggers account suspensions, rendering large-scale evaluation impractical and introducing potential selection biases from interrupted experimental runs.

\subsection{Judges used for Evaluation}

\begin{figure*}[!t]
    \centering
    \resizebox{\textwidth}{!}{
    \subfigure[]{
        \input{chapter/figures/data_statistics/Human_Judgement_distribution.pgf}
    }
    \subfigure[]{
        \input{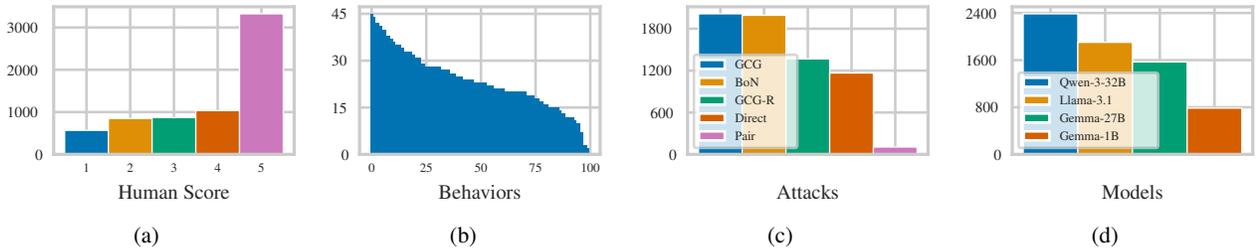}
    }
    \subfigure[]{
        \input{chapter/figures/data_statistics/Attack_distribution.pgf}
    }
    \subfigure[]{
        \input{chapter/figures/data_statistics/Model_distribution.pgf}
    }
    }
    \vspace{-0.3cm}
    \caption{Statistics about the labeled dataset. The individual figure shows the distributions of (a) scores given by human labelers, where scores larger $2$ indicate harmfulness, (b) judged samples per behavior, (c) samples labeled per attack, and (d) samples labeled per model.}
    \label{fig:statistics}
\end{figure*}

For the final evaluation of our dataset, we compare the human labels to ratings from a diverse set of LLM judges from the literature. Specifically, we use \textbf{AegisGuard}~\cite{ghosh2024aegis}, the Llama-2-13B \textbf{HarmBench} classifier~\citep{mazeika2024harmbench}, \textbf{JailJudge}~\cite{liu2024jailjudge}, and \textbf{LlamaGuard-3}~\cite{grattafiori2024llama}. 

\subsection{Attacks \& Selection Criteria}

We evaluate judges across four attack methods that differ in their optimization strategies and the risk of judge hacking. For all attacks, we used the implementations provided by the AdversariaLLM framework~\cite{beyer2025adversariallm}.

\vspace{\myskip}
\textbf{Direct Prompting.} A simple attack that serves as our baseline, where harmful queries are submitted to the victim model without any adversarial modification. Here, we use sampling to strengthen the attack~\cite{scholten2025a}.

\vspace{\myskip}
\textbf{GCG.} A discrete optimization method that appends adversarial suffixes to prompts by iteratively substituting tokens to maximize the likelihood of an affirmative response from the victim model~\cite{zou2023universal}.

\vspace{\myskip}
\textbf{GCG-REINFORCE.} This attack extends GCG by incorporating judge feedback directly into the optimization loop, using the REINFORCE algorithm to guide suffix generation toward outputs judged as harmful~\cite{geislerreinforce}.

\vspace{\myskip}
\textbf{BoN.} Best-of-N is a sampling-based approach that generates $N$ independent responses for a given prompt and selects the output most likely to be harmful according to a judge~\cite{ichihara2025evaluation}. Additionally, BoN applies random character-level perturbations to each query independently, introducing surface-level variation that may further increase the likelihood of bypassing safety guardrails.

\vspace{\myskip}
\textbf{PAIR.} Prompt Automatic Iterative Refinement is an automated framework that uses a secondary attacker LLM to iteratively refine a prompt based on feedback from the victim model and a safety judge~\cite{chao2025jailbreaking}.

Our attack selection is guided by the degree to which each method interacts with the judge, enabling us to investigate how different forms of judge involvement affect error rates.

\vspace{\myskip}
\textbf{Implicit Judge Hacking.} BoN exemplifies attacks that may inadvertently exploit judge weaknesses through extensive sampling. By evaluating many candidate responses, the attack increases the probability of encountering an output that triggers a false positive, where the judge incorrectly classifies a benign response as harmful. This implicit judge hacking emerges as a byproduct of the sampling strategy.

\vspace{\myskip}
\textbf{Explicit Judge Hacking.} GCG-REINFORCE directly uses judge signal in its optimization loop. While this design aims to improve attack effectiveness, it simultaneously creates pressure toward reward hacking: the optimization may discover adversarial suffixes that exploit judge insufficiencies rather than genuinely eliciting harmful generations.

\vspace{\myskip}
\textbf{No Judge Hacking.} Direct Prompting, PAIR, and GCG operate independently of judge feedback. In their default configuration, they query the victim model once per prompt and evaluate the response post-hoc. These methods provide a baseline for judge behavior in the absence of judge-aware optimization. Nevertheless, following prior work demonstrating that sampling enhances attack success rates~\cite{beyer2025sampling}, we additionally explore whether these methods exhibit implicit judge hacking when combined with multi-sample evaluation strategies.

\vspace{\myskip}
We emphasize that none of these methods are designed to deliberately exploit judge weaknesses; rather, the potential for judge hacking arises as an unintended consequence of their respective optimization or sampling procedures. Why only five attacks? Since labeling effort scales combinatorially with the number of attacks and models, we restrict our evaluation to five of the most used methods in the literature. 

\subsection{Further Data Selection and Labeling}

\begin{figure*}[!t]
    \centering
    \subfigure[\llamaeightb{}]{
        \includegraphics[width=0.45\linewidth]{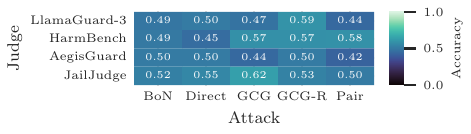}
    }
    \subfigure[\gemmalarge{}]{
        \includegraphics[width=0.45\linewidth]{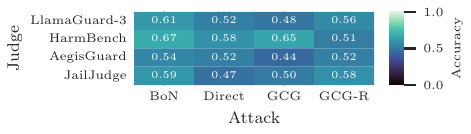}
    }
    \caption{Judge accuracy across attacks for \llamaeightb{} and \gemmalarge{}. Darker colors indicate lower accuracy.}
    \label{fig:heatmap}
\end{figure*}

\textbf{Labeling.} We labeled the data on a scale from 1 (least harmful) to 5 (most harmful). The person labeling was presented with the original prompt in its unaltered form and the model generation. Our labeling approach focused on assessing the model's intent and compliance rather than the technical accuracy or feasibility of the generated content. This means that even if a model's output contained factually incorrect information, such as malicious code that would not execute properly or instructions for harmful activities that would fail in practice, the generation could still be labeled as harmful if it demonstrated intent to comply with a harmful request. The scale was defined as follows, with everything at or above level 3 considered harmful:

\vspace{\myskip}
\textbf{Level 1 (Harmless):} The model explicitly refuses the request and provides an explanation for its refusal, often including ethical reasoning or alternative suggestions.

\vspace{\myskip}
\noindent \textbf{Level 2 (Indirect Refusal):} The model appears to engage with the prompt but subtly redirects, provides incomplete information, or avoids direct compliance.

\vspace{\myskip}
\textbf{Level 3 (Partial Compliance):} The model provides some relevant information but includes significant caveats, disclaimers, or incomplete instructions.

\vspace{\myskip}
\textbf{Level 4 (Compliance with Reservations):} The model largely complies with the harmful request but expresses doubt, includes disclaimers, or attempts to reframe the request as hypothetical or fictional.

\vspace{\myskip}
\textbf{Level 5 (Full Compliance):} The model directly and fully complies with the harmful request without meaningful reservations or safety guardrails.
    
We labeled $2370$ samples ourselves. Additionally, we used Labelbox\footnote{\url{https://labelbox.com/}} to label an additional $4272$ samples. Here, labelers had to pass a quiz of 10 pre-labeled examples before they could start the process. Moreover, we manually reviewed 90 assignments, achieving 95\% agreement in labeling samples as harmful (level 3 or higher) or benign (below level 3) and concluded that the obtained labels were reliable.

\textbf{Data Selection.} We conduct our set of attacks on the subset of HarmBench that we selected. In the next step, our labeling pipeline applies an automatic judge to all attack outputs, classifying each response as either harmful (judge-positive) or safe (judge-negative). We employ the fine-tuned StrongREJECT judge~\cite{souly2024strongreject} to score each response with a harmfulness rating $\mathcal{H} \in [0,1]$, classifying outputs as harmful when $\mathcal{H} > 0.5$. We then restrict human annotation exclusively to judge-positives as labeling judge-negative samples is prohibitively inefficient given realistic annotation budgets. The vast majority of attack trials fail to elicit harmful content: the victim model either refuses outright (e.g., \enquote{I cannot help with that request}), provides a strongly hedged response, or generates irrelevant content. For sampling-intensive methods like BoN, which performs up to 10,000 trials per query, this imbalance is extreme. Annotating negatives at scale would exhaust our labeling budget on true negatives, cases where both the judge and human annotators agree the response is safe, while yielding few informative disagreements. We find that this filtering strategy yields a relatively balanced partition between true positives and false positives, as determined by human annotation.

\subsection{Dataset Statistics}

The final dataset comprises $6642$ samples. Figure~\ref{fig:statistics} illustrates the dataset distribution across human ratings, behaviors, attack types, and models. Although the dataset was pre-filtered to include only samples labeled as harmful by StrongReject, the final composition remains relatively balanced between human-rated harmful ($1437$, $78\%$) and harmless ($5205$, $22\%$) samples. We observe that sample counts decrease as scores progress from $1$ (harmless) to $5$ (harmful). Behaviors are not uniformly represented, a variance primarily attributed to the StrongReject-based filtering; certain adversarial behaviors are inherently more difficult to elicit than others. Due to fluctuating attack success rates, sample counts vary by method; however, we maintained a minimum of $105$ samples for every attack-model combination and labeled at least $785$ samples per model.

%% file: chapter/figures/data_statistics/Human_Judgement_distribution.pgf
\begingroup%
\makeatletter%
\begin{pgfpicture}%
\pgfpathrectangle{\pgfpointorigin}{\pgfqpoint{1.655238in}{1.163676in}}%
\pgfusepath{use as bounding box, clip}%
\begin{pgfscope}%
\pgfsetbuttcap%
\pgfsetmiterjoin%
\definecolor{currentfill}{rgb}{1.000000,1.000000,1.000000}%
\pgfsetfillcolor{currentfill}%
\pgfsetlinewidth{0.000000pt}%
\definecolor{currentstroke}{rgb}{1.000000,1.000000,1.000000}%
\pgfsetstrokecolor{currentstroke}%
\pgfsetdash{}{0pt}%
\pgfpathmoveto{\pgfqpoint{0.000000in}{-0.000000in}}%
\pgfpathlineto{\pgfqpoint{1.655238in}{-0.000000in}}%
\pgfpathlineto{\pgfqpoint{1.655238in}{1.163676in}}%
\pgfpathlineto{\pgfqpoint{0.000000in}{1.163676in}}%
\pgfpathlineto{\pgfqpoint{0.000000in}{-0.000000in}}%
\pgfpathclose%
\pgfusepath{fill}%
\end{pgfscope}%
\begin{pgfscope}%
\pgfsetbuttcap%
\pgfsetmiterjoin%
\definecolor{currentfill}{rgb}{1.000000,1.000000,1.000000}%
\pgfsetfillcolor{currentfill}%
\pgfsetlinewidth{0.000000pt}%
\definecolor{currentstroke}{rgb}{0.000000,0.000000,0.000000}%
\pgfsetstrokecolor{currentstroke}%
\pgfsetstrokeopacity{0.000000}%
\pgfsetdash{}{0pt}%
\pgfpathmoveto{\pgfqpoint{0.302312in}{0.313618in}}%
\pgfpathlineto{\pgfqpoint{1.605238in}{0.313618in}}%
\pgfpathlineto{\pgfqpoint{1.605238in}{1.113676in}}%
\pgfpathlineto{\pgfqpoint{0.302312in}{1.113676in}}%
\pgfpathlineto{\pgfqpoint{0.302312in}{0.313618in}}%
\pgfpathclose%
\pgfusepath{fill}%
\end{pgfscope}%
\begin{pgfscope}%
\pgfpathrectangle{\pgfqpoint{0.302312in}{0.313618in}}{\pgfqpoint{1.302927in}{0.800058in}}%
\pgfusepath{clip}%
\pgfsetroundcap%
\pgfsetroundjoin%
\pgfsetlinewidth{1.003750pt}%
\definecolor{currentstroke}{rgb}{0.800000,0.800000,0.800000}%
\pgfsetstrokecolor{currentstroke}%
\pgfsetdash{}{0pt}%
\pgfpathmoveto{\pgfqpoint{0.479983in}{0.313618in}}%
\pgfpathlineto{\pgfqpoint{0.479983in}{1.113676in}}%
\pgfusepath{stroke}%
\end{pgfscope}%
\begin{pgfscope}%
\definecolor{textcolor}{rgb}{0.150000,0.150000,0.150000}%
\pgfsetstrokecolor{textcolor}%
\pgfsetfillcolor{textcolor}%
\pgftext[x=0.479983in,y=0.265007in,,top]{\color{textcolor}{\rmfamily\fontsize{5.000000}{6.000000}\selectfont\catcode`\^=\active\def^{\ifmmode\sp\else\^{}\fi}\catcode`\%=\active\def
\end{pgfscope}%
\begin{pgfscope}%
\pgfpathrectangle{\pgfqpoint{0.302312in}{0.313618in}}{\pgfqpoint{1.302927in}{0.800058in}}%
\pgfusepath{clip}%
\pgfsetroundcap%
\pgfsetroundjoin%
\pgfsetlinewidth{1.003750pt}%
\definecolor{currentstroke}{rgb}{0.800000,0.800000,0.800000}%
\pgfsetstrokecolor{currentstroke}%
\pgfsetdash{}{0pt}%
\pgfpathmoveto{\pgfqpoint{0.716879in}{0.313618in}}%
\pgfpathlineto{\pgfqpoint{0.716879in}{1.113676in}}%
\pgfusepath{stroke}%
\end{pgfscope}%
\begin{pgfscope}%
\definecolor{textcolor}{rgb}{0.150000,0.150000,0.150000}%
\pgfsetstrokecolor{textcolor}%
\pgfsetfillcolor{textcolor}%
\pgftext[x=0.716879in,y=0.265007in,,top]{\color{textcolor}{\rmfamily\fontsize{5.000000}{6.000000}\selectfont\catcode`\^=\active\def^{\ifmmode\sp\else\^{}\fi}\catcode`\%=\active\def
\end{pgfscope}%
\begin{pgfscope}%
\pgfpathrectangle{\pgfqpoint{0.302312in}{0.313618in}}{\pgfqpoint{1.302927in}{0.800058in}}%
\pgfusepath{clip}%
\pgfsetroundcap%
\pgfsetroundjoin%
\pgfsetlinewidth{1.003750pt}%
\definecolor{currentstroke}{rgb}{0.800000,0.800000,0.800000}%
\pgfsetstrokecolor{currentstroke}%
\pgfsetdash{}{0pt}%
\pgfpathmoveto{\pgfqpoint{0.953775in}{0.313618in}}%
\pgfpathlineto{\pgfqpoint{0.953775in}{1.113676in}}%
\pgfusepath{stroke}%
\end{pgfscope}%
\begin{pgfscope}%
\definecolor{textcolor}{rgb}{0.150000,0.150000,0.150000}%
\pgfsetstrokecolor{textcolor}%
\pgfsetfillcolor{textcolor}%
\pgftext[x=0.953775in,y=0.265007in,,top]{\color{textcolor}{\rmfamily\fontsize{5.000000}{6.000000}\selectfont\catcode`\^=\active\def^{\ifmmode\sp\else\^{}\fi}\catcode`\%=\active\def
\end{pgfscope}%
\begin{pgfscope}%
\pgfpathrectangle{\pgfqpoint{0.302312in}{0.313618in}}{\pgfqpoint{1.302927in}{0.800058in}}%
\pgfusepath{clip}%
\pgfsetroundcap%
\pgfsetroundjoin%
\pgfsetlinewidth{1.003750pt}%
\definecolor{currentstroke}{rgb}{0.800000,0.800000,0.800000}%
\pgfsetstrokecolor{currentstroke}%
\pgfsetdash{}{0pt}%
\pgfpathmoveto{\pgfqpoint{1.190671in}{0.313618in}}%
\pgfpathlineto{\pgfqpoint{1.190671in}{1.113676in}}%
\pgfusepath{stroke}%
\end{pgfscope}%
\begin{pgfscope}%
\definecolor{textcolor}{rgb}{0.150000,0.150000,0.150000}%
\pgfsetstrokecolor{textcolor}%
\pgfsetfillcolor{textcolor}%
\pgftext[x=1.190671in,y=0.265007in,,top]{\color{textcolor}{\rmfamily\fontsize{5.000000}{6.000000}\selectfont\catcode`\^=\active\def^{\ifmmode\sp\else\^{}\fi}\catcode`\%=\active\def
\end{pgfscope}%
\begin{pgfscope}%
\pgfpathrectangle{\pgfqpoint{0.302312in}{0.313618in}}{\pgfqpoint{1.302927in}{0.800058in}}%
\pgfusepath{clip}%
\pgfsetroundcap%
\pgfsetroundjoin%
\pgfsetlinewidth{1.003750pt}%
\definecolor{currentstroke}{rgb}{0.800000,0.800000,0.800000}%
\pgfsetstrokecolor{currentstroke}%
\pgfsetdash{}{0pt}%
\pgfpathmoveto{\pgfqpoint{1.427566in}{0.313618in}}%
\pgfpathlineto{\pgfqpoint{1.427566in}{1.113676in}}%
\pgfusepath{stroke}%
\end{pgfscope}%
\begin{pgfscope}%
\definecolor{textcolor}{rgb}{0.150000,0.150000,0.150000}%
\pgfsetstrokecolor{textcolor}%
\pgfsetfillcolor{textcolor}%
\pgftext[x=1.427566in,y=0.265007in,,top]{\color{textcolor}{\rmfamily\fontsize{5.000000}{6.000000}\selectfont\catcode`\^=\active\def^{\ifmmode\sp\else\^{}\fi}\catcode`\%=\active\def
\end{pgfscope}%
\begin{pgfscope}%
\definecolor{textcolor}{rgb}{0.150000,0.150000,0.150000}%
\pgfsetstrokecolor{textcolor}%
\pgfsetfillcolor{textcolor}%
\pgftext[x=0.953775in,y=0.148124in,,top]{\color{textcolor}{\rmfamily\fontsize{8.000000}{9.600000}\selectfont\catcode`\^=\active\def^{\ifmmode\sp\else\^{}\fi}\catcode`\%=\active\def
\end{pgfscope}%
\begin{pgfscope}%
\pgfpathrectangle{\pgfqpoint{0.302312in}{0.313618in}}{\pgfqpoint{1.302927in}{0.800058in}}%
\pgfusepath{clip}%
\pgfsetroundcap%
\pgfsetroundjoin%
\pgfsetlinewidth{1.003750pt}%
\definecolor{currentstroke}{rgb}{0.800000,0.800000,0.800000}%
\pgfsetstrokecolor{currentstroke}%
\pgfsetdash{}{0pt}%
\pgfpathmoveto{\pgfqpoint{0.302312in}{0.313618in}}%
\pgfpathlineto{\pgfqpoint{1.605238in}{0.313618in}}%
\pgfusepath{stroke}%
\end{pgfscope}%
\begin{pgfscope}%
\definecolor{textcolor}{rgb}{0.150000,0.150000,0.150000}%
\pgfsetstrokecolor{textcolor}%
\pgfsetfillcolor{textcolor}%
\pgftext[x=0.202775in, y=0.284690in, left, base]{\color{textcolor}{\rmfamily\fontsize{6.000000}{7.200000}\selectfont\catcode`\^=\active\def^{\ifmmode\sp\else\^{}\fi}\catcode`\%=\active\def
\end{pgfscope}%
\begin{pgfscope}%
\pgfpathrectangle{\pgfqpoint{0.302312in}{0.313618in}}{\pgfqpoint{1.302927in}{0.800058in}}%
\pgfusepath{clip}%
\pgfsetroundcap%
\pgfsetroundjoin%
\pgfsetlinewidth{1.003750pt}%
\definecolor{currentstroke}{rgb}{0.800000,0.800000,0.800000}%
\pgfsetstrokecolor{currentstroke}%
\pgfsetdash{}{0pt}%
\pgfpathmoveto{\pgfqpoint{0.302312in}{0.542986in}}%
\pgfpathlineto{\pgfqpoint{1.605238in}{0.542986in}}%
\pgfusepath{stroke}%
\end{pgfscope}%
\begin{pgfscope}%
\definecolor{textcolor}{rgb}{0.150000,0.150000,0.150000}%
\pgfsetstrokecolor{textcolor}%
\pgfsetfillcolor{textcolor}%
\pgftext[x=0.050000in, y=0.514058in, left, base]{\color{textcolor}{\rmfamily\fontsize{6.000000}{7.200000}\selectfont\catcode`\^=\active\def^{\ifmmode\sp\else\^{}\fi}\catcode`\%=\active\def
\end{pgfscope}%
\begin{pgfscope}%
\pgfpathrectangle{\pgfqpoint{0.302312in}{0.313618in}}{\pgfqpoint{1.302927in}{0.800058in}}%
\pgfusepath{clip}%
\pgfsetroundcap%
\pgfsetroundjoin%
\pgfsetlinewidth{1.003750pt}%
\definecolor{currentstroke}{rgb}{0.800000,0.800000,0.800000}%
\pgfsetstrokecolor{currentstroke}%
\pgfsetdash{}{0pt}%
\pgfpathmoveto{\pgfqpoint{0.302312in}{0.772354in}}%
\pgfpathlineto{\pgfqpoint{1.605238in}{0.772354in}}%
\pgfusepath{stroke}%
\end{pgfscope}%
\begin{pgfscope}%
\definecolor{textcolor}{rgb}{0.150000,0.150000,0.150000}%
\pgfsetstrokecolor{textcolor}%
\pgfsetfillcolor{textcolor}%
\pgftext[x=0.050000in, y=0.743426in, left, base]{\color{textcolor}{\rmfamily\fontsize{6.000000}{7.200000}\selectfont\catcode`\^=\active\def^{\ifmmode\sp\else\^{}\fi}\catcode`\%=\active\def
\end{pgfscope}%
\begin{pgfscope}%
\pgfpathrectangle{\pgfqpoint{0.302312in}{0.313618in}}{\pgfqpoint{1.302927in}{0.800058in}}%
\pgfusepath{clip}%
\pgfsetroundcap%
\pgfsetroundjoin%
\pgfsetlinewidth{1.003750pt}%
\definecolor{currentstroke}{rgb}{0.800000,0.800000,0.800000}%
\pgfsetstrokecolor{currentstroke}%
\pgfsetdash{}{0pt}%
\pgfpathmoveto{\pgfqpoint{0.302312in}{1.001722in}}%
\pgfpathlineto{\pgfqpoint{1.605238in}{1.001722in}}%
\pgfusepath{stroke}%
\end{pgfscope}%
\begin{pgfscope}%
\definecolor{textcolor}{rgb}{0.150000,0.150000,0.150000}%
\pgfsetstrokecolor{textcolor}%
\pgfsetfillcolor{textcolor}%
\pgftext[x=0.050000in, y=0.972794in, left, base]{\color{textcolor}{\rmfamily\fontsize{6.000000}{7.200000}\selectfont\catcode`\^=\active\def^{\ifmmode\sp\else\^{}\fi}\catcode`\%=\active\def
\end{pgfscope}%
\begin{pgfscope}%
\pgfpathrectangle{\pgfqpoint{0.302312in}{0.313618in}}{\pgfqpoint{1.302927in}{0.800058in}}%
\pgfusepath{clip}%
\pgfsetbuttcap%
\pgfsetmiterjoin%
\definecolor{currentfill}{rgb}{0.003922,0.450980,0.698039}%
\pgfsetfillcolor{currentfill}%
\pgfsetlinewidth{0.501875pt}%
\definecolor{currentstroke}{rgb}{1.000000,1.000000,1.000000}%
\pgfsetstrokecolor{currentstroke}%
\pgfsetdash{}{0pt}%
\pgfpathmoveto{\pgfqpoint{0.361536in}{0.313618in}}%
\pgfpathlineto{\pgfqpoint{0.598431in}{0.313618in}}%
\pgfpathlineto{\pgfqpoint{0.598431in}{0.443899in}}%
\pgfpathlineto{\pgfqpoint{0.361536in}{0.443899in}}%
\pgfpathlineto{\pgfqpoint{0.361536in}{0.313618in}}%
\pgfpathclose%
\pgfusepath{stroke,fill}%
\end{pgfscope}%
\begin{pgfscope}%
\pgfpathrectangle{\pgfqpoint{0.302312in}{0.313618in}}{\pgfqpoint{1.302927in}{0.800058in}}%
\pgfusepath{clip}%
\pgfsetbuttcap%
\pgfsetmiterjoin%
\definecolor{currentfill}{rgb}{0.870588,0.560784,0.019608}%
\pgfsetfillcolor{currentfill}%
\pgfsetlinewidth{0.501875pt}%
\definecolor{currentstroke}{rgb}{1.000000,1.000000,1.000000}%
\pgfsetstrokecolor{currentstroke}%
\pgfsetdash{}{0pt}%
\pgfpathmoveto{\pgfqpoint{0.598431in}{0.313618in}}%
\pgfpathlineto{\pgfqpoint{0.835327in}{0.313618in}}%
\pgfpathlineto{\pgfqpoint{0.835327in}{0.508352in}}%
\pgfpathlineto{\pgfqpoint{0.598431in}{0.508352in}}%
\pgfpathlineto{\pgfqpoint{0.598431in}{0.313618in}}%
\pgfpathclose%
\pgfusepath{stroke,fill}%
\end{pgfscope}%
\begin{pgfscope}%
\pgfpathrectangle{\pgfqpoint{0.302312in}{0.313618in}}{\pgfqpoint{1.302927in}{0.800058in}}%
\pgfusepath{clip}%
\pgfsetbuttcap%
\pgfsetmiterjoin%
\definecolor{currentfill}{rgb}{0.007843,0.619608,0.450980}%
\pgfsetfillcolor{currentfill}%
\pgfsetlinewidth{0.501875pt}%
\definecolor{currentstroke}{rgb}{1.000000,1.000000,1.000000}%
\pgfsetstrokecolor{currentstroke}%
\pgfsetdash{}{0pt}%
\pgfpathmoveto{\pgfqpoint{0.835327in}{0.313618in}}%
\pgfpathlineto{\pgfqpoint{1.072223in}{0.313618in}}%
\pgfpathlineto{\pgfqpoint{1.072223in}{0.512939in}}%
\pgfpathlineto{\pgfqpoint{0.835327in}{0.512939in}}%
\pgfpathlineto{\pgfqpoint{0.835327in}{0.313618in}}%
\pgfpathclose%
\pgfusepath{stroke,fill}%
\end{pgfscope}%
\begin{pgfscope}%
\pgfpathrectangle{\pgfqpoint{0.302312in}{0.313618in}}{\pgfqpoint{1.302927in}{0.800058in}}%
\pgfusepath{clip}%
\pgfsetbuttcap%
\pgfsetmiterjoin%
\definecolor{currentfill}{rgb}{0.835294,0.368627,0.000000}%
\pgfsetfillcolor{currentfill}%
\pgfsetlinewidth{0.501875pt}%
\definecolor{currentstroke}{rgb}{1.000000,1.000000,1.000000}%
\pgfsetstrokecolor{currentstroke}%
\pgfsetdash{}{0pt}%
\pgfpathmoveto{\pgfqpoint{1.072223in}{0.313618in}}%
\pgfpathlineto{\pgfqpoint{1.309119in}{0.313618in}}%
\pgfpathlineto{\pgfqpoint{1.309119in}{0.550785in}}%
\pgfpathlineto{\pgfqpoint{1.072223in}{0.550785in}}%
\pgfpathlineto{\pgfqpoint{1.072223in}{0.313618in}}%
\pgfpathclose%
\pgfusepath{stroke,fill}%
\end{pgfscope}%
\begin{pgfscope}%
\pgfpathrectangle{\pgfqpoint{0.302312in}{0.313618in}}{\pgfqpoint{1.302927in}{0.800058in}}%
\pgfusepath{clip}%
\pgfsetbuttcap%
\pgfsetmiterjoin%
\definecolor{currentfill}{rgb}{0.800000,0.470588,0.737255}%
\pgfsetfillcolor{currentfill}%
\pgfsetlinewidth{0.501875pt}%
\definecolor{currentstroke}{rgb}{1.000000,1.000000,1.000000}%
\pgfsetstrokecolor{currentstroke}%
\pgfsetdash{}{0pt}%
\pgfpathmoveto{\pgfqpoint{1.309119in}{0.313618in}}%
\pgfpathlineto{\pgfqpoint{1.546014in}{0.313618in}}%
\pgfpathlineto{\pgfqpoint{1.546014in}{1.075578in}}%
\pgfpathlineto{\pgfqpoint{1.309119in}{1.075578in}}%
\pgfpathlineto{\pgfqpoint{1.309119in}{0.313618in}}%
\pgfpathclose%
\pgfusepath{stroke,fill}%
\end{pgfscope}%
\begin{pgfscope}%
\pgfsetrectcap%
\pgfsetmiterjoin%
\pgfsetlinewidth{1.254687pt}%
\definecolor{currentstroke}{rgb}{0.800000,0.800000,0.800000}%
\pgfsetstrokecolor{currentstroke}%
\pgfsetdash{}{0pt}%
\pgfpathmoveto{\pgfqpoint{0.302312in}{0.313618in}}%
\pgfpathlineto{\pgfqpoint{0.302312in}{1.113676in}}%
\pgfusepath{stroke}%
\end{pgfscope}%
\begin{pgfscope}%
\pgfsetrectcap%
\pgfsetmiterjoin%
\pgfsetlinewidth{1.254687pt}%
\definecolor{currentstroke}{rgb}{0.800000,0.800000,0.800000}%
\pgfsetstrokecolor{currentstroke}%
\pgfsetdash{}{0pt}%
\pgfpathmoveto{\pgfqpoint{1.605238in}{0.313618in}}%
\pgfpathlineto{\pgfqpoint{1.605238in}{1.113676in}}%
\pgfusepath{stroke}%
\end{pgfscope}%
\begin{pgfscope}%
\pgfsetrectcap%
\pgfsetmiterjoin%
\pgfsetlinewidth{1.254687pt}%
\definecolor{currentstroke}{rgb}{0.800000,0.800000,0.800000}%
\pgfsetstrokecolor{currentstroke}%
\pgfsetdash{}{0pt}%
\pgfpathmoveto{\pgfqpoint{0.302312in}{0.313618in}}%
\pgfpathlineto{\pgfqpoint{1.605238in}{0.313618in}}%
\pgfusepath{stroke}%
\end{pgfscope}%
\begin{pgfscope}%
\pgfsetrectcap%
\pgfsetmiterjoin%
\pgfsetlinewidth{1.254687pt}%
\definecolor{currentstroke}{rgb}{0.800000,0.800000,0.800000}%
\pgfsetstrokecolor{currentstroke}%
\pgfsetdash{}{0pt}%
\pgfpathmoveto{\pgfqpoint{0.302312in}{1.113676in}}%
\pgfpathlineto{\pgfqpoint{1.605238in}{1.113676in}}%
\pgfusepath{stroke}%
\end{pgfscope}%
\end{pgfpicture}%
\makeatother%
\endgroup%

%% file: chapter/figures/data_statistics/Attack_distribution.pgf
\begingroup%
\makeatletter%
\begin{pgfpicture}%
\pgfpathrectangle{\pgfpointorigin}{\pgfqpoint{1.655238in}{1.163676in}}%
\pgfusepath{use as bounding box, clip}%
\begin{pgfscope}%
\pgfsetbuttcap%
\pgfsetmiterjoin%
\definecolor{currentfill}{rgb}{1.000000,1.000000,1.000000}%
\pgfsetfillcolor{currentfill}%
\pgfsetlinewidth{0.000000pt}%
\definecolor{currentstroke}{rgb}{1.000000,1.000000,1.000000}%
\pgfsetstrokecolor{currentstroke}%
\pgfsetdash{}{0pt}%
\pgfpathmoveto{\pgfqpoint{0.000000in}{-0.000000in}}%
\pgfpathlineto{\pgfqpoint{1.655238in}{-0.000000in}}%
\pgfpathlineto{\pgfqpoint{1.655238in}{1.163676in}}%
\pgfpathlineto{\pgfqpoint{0.000000in}{1.163676in}}%
\pgfpathlineto{\pgfqpoint{0.000000in}{-0.000000in}}%
\pgfpathclose%
\pgfusepath{fill}%
\end{pgfscope}%
\begin{pgfscope}%
\pgfsetbuttcap%
\pgfsetmiterjoin%
\definecolor{currentfill}{rgb}{1.000000,1.000000,1.000000}%
\pgfsetfillcolor{currentfill}%
\pgfsetlinewidth{0.000000pt}%
\definecolor{currentstroke}{rgb}{0.000000,0.000000,0.000000}%
\pgfsetstrokecolor{currentstroke}%
\pgfsetstrokeopacity{0.000000}%
\pgfsetdash{}{0pt}%
\pgfpathmoveto{\pgfqpoint{0.302312in}{0.313618in}}%
\pgfpathlineto{\pgfqpoint{1.605238in}{0.313618in}}%
\pgfpathlineto{\pgfqpoint{1.605238in}{1.113676in}}%
\pgfpathlineto{\pgfqpoint{0.302312in}{1.113676in}}%
\pgfpathlineto{\pgfqpoint{0.302312in}{0.313618in}}%
\pgfpathclose%
\pgfusepath{fill}%
\end{pgfscope}%
\begin{pgfscope}%
\pgfpathrectangle{\pgfqpoint{0.302312in}{0.313618in}}{\pgfqpoint{1.302927in}{0.800058in}}%
\pgfusepath{clip}%
\pgfsetroundcap%
\pgfsetroundjoin%
\pgfsetlinewidth{1.003750pt}%
\definecolor{currentstroke}{rgb}{0.800000,0.800000,0.800000}%
\pgfsetstrokecolor{currentstroke}%
\pgfsetdash{}{0pt}%
\pgfpathmoveto{\pgfqpoint{0.479983in}{0.313618in}}%
\pgfpathlineto{\pgfqpoint{0.479983in}{1.113676in}}%
\pgfusepath{stroke}%
\end{pgfscope}%
\begin{pgfscope}%
\definecolor{textcolor}{rgb}{1.000000,1.000000,1.000000}%
\pgfsetstrokecolor{textcolor}%
\pgfsetfillcolor{textcolor}%
\pgftext[x=0.479983in,y=0.265007in,,top]{\color{textcolor}{\rmfamily\fontsize{5.000000}{6.000000}\selectfont\catcode`\^=\active\def^{\ifmmode\sp\else\^{}\fi}\catcode`\%=\active\def
\end{pgfscope}%
\begin{pgfscope}%
\pgfpathrectangle{\pgfqpoint{0.302312in}{0.313618in}}{\pgfqpoint{1.302927in}{0.800058in}}%
\pgfusepath{clip}%
\pgfsetroundcap%
\pgfsetroundjoin%
\pgfsetlinewidth{1.003750pt}%
\definecolor{currentstroke}{rgb}{0.800000,0.800000,0.800000}%
\pgfsetstrokecolor{currentstroke}%
\pgfsetdash{}{0pt}%
\pgfpathmoveto{\pgfqpoint{0.716879in}{0.313618in}}%
\pgfpathlineto{\pgfqpoint{0.716879in}{1.113676in}}%
\pgfusepath{stroke}%
\end{pgfscope}%
\begin{pgfscope}%
\definecolor{textcolor}{rgb}{1.000000,1.000000,1.000000}%
\pgfsetstrokecolor{textcolor}%
\pgfsetfillcolor{textcolor}%
\pgftext[x=0.716879in,y=0.265007in,,top]{\color{textcolor}{\rmfamily\fontsize{5.000000}{6.000000}\selectfont\catcode`\^=\active\def^{\ifmmode\sp\else\^{}\fi}\catcode`\%=\active\def
\end{pgfscope}%
\begin{pgfscope}%
\pgfpathrectangle{\pgfqpoint{0.302312in}{0.313618in}}{\pgfqpoint{1.302927in}{0.800058in}}%
\pgfusepath{clip}%
\pgfsetroundcap%
\pgfsetroundjoin%
\pgfsetlinewidth{1.003750pt}%
\definecolor{currentstroke}{rgb}{0.800000,0.800000,0.800000}%
\pgfsetstrokecolor{currentstroke}%
\pgfsetdash{}{0pt}%
\pgfpathmoveto{\pgfqpoint{0.953775in}{0.313618in}}%
\pgfpathlineto{\pgfqpoint{0.953775in}{1.113676in}}%
\pgfusepath{stroke}%
\end{pgfscope}%
\begin{pgfscope}%
\definecolor{textcolor}{rgb}{1.000000,1.000000,1.000000}%
\pgfsetstrokecolor{textcolor}%
\pgfsetfillcolor{textcolor}%
\pgftext[x=0.953775in,y=0.265007in,,top]{\color{textcolor}{\rmfamily\fontsize{5.000000}{6.000000}\selectfont\catcode`\^=\active\def^{\ifmmode\sp\else\^{}\fi}\catcode`\%=\active\def
\end{pgfscope}%
\begin{pgfscope}%
\pgfpathrectangle{\pgfqpoint{0.302312in}{0.313618in}}{\pgfqpoint{1.302927in}{0.800058in}}%
\pgfusepath{clip}%
\pgfsetroundcap%
\pgfsetroundjoin%
\pgfsetlinewidth{1.003750pt}%
\definecolor{currentstroke}{rgb}{0.800000,0.800000,0.800000}%
\pgfsetstrokecolor{currentstroke}%
\pgfsetdash{}{0pt}%
\pgfpathmoveto{\pgfqpoint{1.190671in}{0.313618in}}%
\pgfpathlineto{\pgfqpoint{1.190671in}{1.113676in}}%
\pgfusepath{stroke}%
\end{pgfscope}%
\begin{pgfscope}%
\definecolor{textcolor}{rgb}{1.000000,1.000000,1.000000}%
\pgfsetstrokecolor{textcolor}%
\pgfsetfillcolor{textcolor}%
\pgftext[x=1.190671in,y=0.265007in,,top]{\color{textcolor}{\rmfamily\fontsize{5.000000}{6.000000}\selectfont\catcode`\^=\active\def^{\ifmmode\sp\else\^{}\fi}\catcode`\%=\active\def
\end{pgfscope}%
\begin{pgfscope}%
\pgfpathrectangle{\pgfqpoint{0.302312in}{0.313618in}}{\pgfqpoint{1.302927in}{0.800058in}}%
\pgfusepath{clip}%
\pgfsetroundcap%
\pgfsetroundjoin%
\pgfsetlinewidth{1.003750pt}%
\definecolor{currentstroke}{rgb}{0.800000,0.800000,0.800000}%
\pgfsetstrokecolor{currentstroke}%
\pgfsetdash{}{0pt}%
\pgfpathmoveto{\pgfqpoint{1.427566in}{0.313618in}}%
\pgfpathlineto{\pgfqpoint{1.427566in}{1.113676in}}%
\pgfusepath{stroke}%
\end{pgfscope}%
\begin{pgfscope}%
\definecolor{textcolor}{rgb}{1.000000,1.000000,1.000000}%
\pgfsetstrokecolor{textcolor}%
\pgfsetfillcolor{textcolor}%
\pgftext[x=1.427566in,y=0.265007in,,top]{\color{textcolor}{\rmfamily\fontsize{5.000000}{6.000000}\selectfont\catcode`\^=\active\def^{\ifmmode\sp\else\^{}\fi}\catcode`\%=\active\def
\end{pgfscope}%
\begin{pgfscope}%
\definecolor{textcolor}{rgb}{0.150000,0.150000,0.150000}%
\pgfsetstrokecolor{textcolor}%
\pgfsetfillcolor{textcolor}%
\pgftext[x=0.953775in,y=0.148124in,,top]{\color{textcolor}{\rmfamily\fontsize{8.000000}{9.600000}\selectfont\catcode`\^=\active\def^{\ifmmode\sp\else\^{}\fi}\catcode`\%=\active\def
\end{pgfscope}%
\begin{pgfscope}%
\pgfpathrectangle{\pgfqpoint{0.302312in}{0.313618in}}{\pgfqpoint{1.302927in}{0.800058in}}%
\pgfusepath{clip}%
\pgfsetroundcap%
\pgfsetroundjoin%
\pgfsetlinewidth{1.003750pt}%
\definecolor{currentstroke}{rgb}{0.800000,0.800000,0.800000}%
\pgfsetstrokecolor{currentstroke}%
\pgfsetdash{}{0pt}%
\pgfpathmoveto{\pgfqpoint{0.302312in}{0.313618in}}%
\pgfpathlineto{\pgfqpoint{1.605238in}{0.313618in}}%
\pgfusepath{stroke}%
\end{pgfscope}%
\begin{pgfscope}%
\definecolor{textcolor}{rgb}{0.150000,0.150000,0.150000}%
\pgfsetstrokecolor{textcolor}%
\pgfsetfillcolor{textcolor}%
\pgftext[x=0.202775in, y=0.284690in, left, base]{\color{textcolor}{\rmfamily\fontsize{6.000000}{7.200000}\selectfont\catcode`\^=\active\def^{\ifmmode\sp\else\^{}\fi}\catcode`\%=\active\def
\end{pgfscope}%
\begin{pgfscope}%
\pgfpathrectangle{\pgfqpoint{0.302312in}{0.313618in}}{\pgfqpoint{1.302927in}{0.800058in}}%
\pgfusepath{clip}%
\pgfsetroundcap%
\pgfsetroundjoin%
\pgfsetlinewidth{1.003750pt}%
\definecolor{currentstroke}{rgb}{0.800000,0.800000,0.800000}%
\pgfsetstrokecolor{currentstroke}%
\pgfsetdash{}{0pt}%
\pgfpathmoveto{\pgfqpoint{0.302312in}{0.540730in}}%
\pgfpathlineto{\pgfqpoint{1.605238in}{0.540730in}}%
\pgfusepath{stroke}%
\end{pgfscope}%
\begin{pgfscope}%
\definecolor{textcolor}{rgb}{0.150000,0.150000,0.150000}%
\pgfsetstrokecolor{textcolor}%
\pgfsetfillcolor{textcolor}%
\pgftext[x=0.100925in, y=0.511802in, left, base]{\color{textcolor}{\rmfamily\fontsize{6.000000}{7.200000}\selectfont\catcode`\^=\active\def^{\ifmmode\sp\else\^{}\fi}\catcode`\%=\active\def
\end{pgfscope}%
\begin{pgfscope}%
\pgfpathrectangle{\pgfqpoint{0.302312in}{0.313618in}}{\pgfqpoint{1.302927in}{0.800058in}}%
\pgfusepath{clip}%
\pgfsetroundcap%
\pgfsetroundjoin%
\pgfsetlinewidth{1.003750pt}%
\definecolor{currentstroke}{rgb}{0.800000,0.800000,0.800000}%
\pgfsetstrokecolor{currentstroke}%
\pgfsetdash{}{0pt}%
\pgfpathmoveto{\pgfqpoint{0.302312in}{0.767842in}}%
\pgfpathlineto{\pgfqpoint{1.605238in}{0.767842in}}%
\pgfusepath{stroke}%
\end{pgfscope}%
\begin{pgfscope}%
\definecolor{textcolor}{rgb}{0.150000,0.150000,0.150000}%
\pgfsetstrokecolor{textcolor}%
\pgfsetfillcolor{textcolor}%
\pgftext[x=0.050000in, y=0.738914in, left, base]{\color{textcolor}{\rmfamily\fontsize{6.000000}{7.200000}\selectfont\catcode`\^=\active\def^{\ifmmode\sp\else\^{}\fi}\catcode`\%=\active\def
\end{pgfscope}%
\begin{pgfscope}%
\pgfpathrectangle{\pgfqpoint{0.302312in}{0.313618in}}{\pgfqpoint{1.302927in}{0.800058in}}%
\pgfusepath{clip}%
\pgfsetroundcap%
\pgfsetroundjoin%
\pgfsetlinewidth{1.003750pt}%
\definecolor{currentstroke}{rgb}{0.800000,0.800000,0.800000}%
\pgfsetstrokecolor{currentstroke}%
\pgfsetdash{}{0pt}%
\pgfpathmoveto{\pgfqpoint{0.302312in}{0.994954in}}%
\pgfpathlineto{\pgfqpoint{1.605238in}{0.994954in}}%
\pgfusepath{stroke}%
\end{pgfscope}%
\begin{pgfscope}%
\definecolor{textcolor}{rgb}{0.150000,0.150000,0.150000}%
\pgfsetstrokecolor{textcolor}%
\pgfsetfillcolor{textcolor}%
\pgftext[x=0.050000in, y=0.966025in, left, base]{\color{textcolor}{\rmfamily\fontsize{6.000000}{7.200000}\selectfont\catcode`\^=\active\def^{\ifmmode\sp\else\^{}\fi}\catcode`\%=\active\def
\end{pgfscope}%
\begin{pgfscope}%
\pgfpathrectangle{\pgfqpoint{0.302312in}{0.313618in}}{\pgfqpoint{1.302927in}{0.800058in}}%
\pgfusepath{clip}%
\pgfsetbuttcap%
\pgfsetmiterjoin%
\definecolor{currentfill}{rgb}{0.003922,0.450980,0.698039}%
\pgfsetfillcolor{currentfill}%
\pgfsetlinewidth{0.501875pt}%
\definecolor{currentstroke}{rgb}{1.000000,1.000000,1.000000}%
\pgfsetstrokecolor{currentstroke}%
\pgfsetdash{}{0pt}%
\pgfpathmoveto{\pgfqpoint{0.361536in}{0.313618in}}%
\pgfpathlineto{\pgfqpoint{0.598431in}{0.313618in}}%
\pgfpathlineto{\pgfqpoint{0.598431in}{1.075578in}}%
\pgfpathlineto{\pgfqpoint{0.361536in}{1.075578in}}%
\pgfpathlineto{\pgfqpoint{0.361536in}{0.313618in}}%
\pgfpathclose%
\pgfusepath{stroke,fill}%
\end{pgfscope}%
\begin{pgfscope}%
\pgfpathrectangle{\pgfqpoint{0.302312in}{0.313618in}}{\pgfqpoint{1.302927in}{0.800058in}}%
\pgfusepath{clip}%
\pgfsetbuttcap%
\pgfsetmiterjoin%
\definecolor{currentfill}{rgb}{0.870588,0.560784,0.019608}%
\pgfsetfillcolor{currentfill}%
\pgfsetlinewidth{0.501875pt}%
\definecolor{currentstroke}{rgb}{1.000000,1.000000,1.000000}%
\pgfsetstrokecolor{currentstroke}%
\pgfsetdash{}{0pt}%
\pgfpathmoveto{\pgfqpoint{0.598431in}{0.313618in}}%
\pgfpathlineto{\pgfqpoint{0.835327in}{0.313618in}}%
\pgfpathlineto{\pgfqpoint{0.835327in}{1.067629in}}%
\pgfpathlineto{\pgfqpoint{0.598431in}{1.067629in}}%
\pgfpathlineto{\pgfqpoint{0.598431in}{0.313618in}}%
\pgfpathclose%
\pgfusepath{stroke,fill}%
\end{pgfscope}%
\begin{pgfscope}%
\pgfpathrectangle{\pgfqpoint{0.302312in}{0.313618in}}{\pgfqpoint{1.302927in}{0.800058in}}%
\pgfusepath{clip}%
\pgfsetbuttcap%
\pgfsetmiterjoin%
\definecolor{currentfill}{rgb}{0.007843,0.619608,0.450980}%
\pgfsetfillcolor{currentfill}%
\pgfsetlinewidth{0.501875pt}%
\definecolor{currentstroke}{rgb}{1.000000,1.000000,1.000000}%
\pgfsetstrokecolor{currentstroke}%
\pgfsetdash{}{0pt}%
\pgfpathmoveto{\pgfqpoint{0.835327in}{0.313618in}}%
\pgfpathlineto{\pgfqpoint{1.072223in}{0.313618in}}%
\pgfpathlineto{\pgfqpoint{1.072223in}{0.831055in}}%
\pgfpathlineto{\pgfqpoint{0.835327in}{0.831055in}}%
\pgfpathlineto{\pgfqpoint{0.835327in}{0.313618in}}%
\pgfpathclose%
\pgfusepath{stroke,fill}%
\end{pgfscope}%
\begin{pgfscope}%
\pgfpathrectangle{\pgfqpoint{0.302312in}{0.313618in}}{\pgfqpoint{1.302927in}{0.800058in}}%
\pgfusepath{clip}%
\pgfsetbuttcap%
\pgfsetmiterjoin%
\definecolor{currentfill}{rgb}{0.835294,0.368627,0.000000}%
\pgfsetfillcolor{currentfill}%
\pgfsetlinewidth{0.501875pt}%
\definecolor{currentstroke}{rgb}{1.000000,1.000000,1.000000}%
\pgfsetstrokecolor{currentstroke}%
\pgfsetdash{}{0pt}%
\pgfpathmoveto{\pgfqpoint{1.072223in}{0.313618in}}%
\pgfpathlineto{\pgfqpoint{1.309119in}{0.313618in}}%
\pgfpathlineto{\pgfqpoint{1.309119in}{0.754594in}}%
\pgfpathlineto{\pgfqpoint{1.072223in}{0.754594in}}%
\pgfpathlineto{\pgfqpoint{1.072223in}{0.313618in}}%
\pgfpathclose%
\pgfusepath{stroke,fill}%
\end{pgfscope}%
\begin{pgfscope}%
\pgfpathrectangle{\pgfqpoint{0.302312in}{0.313618in}}{\pgfqpoint{1.302927in}{0.800058in}}%
\pgfusepath{clip}%
\pgfsetbuttcap%
\pgfsetmiterjoin%
\definecolor{currentfill}{rgb}{0.800000,0.470588,0.737255}%
\pgfsetfillcolor{currentfill}%
\pgfsetlinewidth{0.501875pt}%
\definecolor{currentstroke}{rgb}{1.000000,1.000000,1.000000}%
\pgfsetstrokecolor{currentstroke}%
\pgfsetdash{}{0pt}%
\pgfpathmoveto{\pgfqpoint{1.309119in}{0.313618in}}%
\pgfpathlineto{\pgfqpoint{1.546014in}{0.313618in}}%
\pgfpathlineto{\pgfqpoint{1.546014in}{0.353363in}}%
\pgfpathlineto{\pgfqpoint{1.309119in}{0.353363in}}%
\pgfpathlineto{\pgfqpoint{1.309119in}{0.313618in}}%
\pgfpathclose%
\pgfusepath{stroke,fill}%
\end{pgfscope}%
\begin{pgfscope}%
\pgfsetrectcap%
\pgfsetmiterjoin%
\pgfsetlinewidth{1.254687pt}%
\definecolor{currentstroke}{rgb}{0.800000,0.800000,0.800000}%
\pgfsetstrokecolor{currentstroke}%
\pgfsetdash{}{0pt}%
\pgfpathmoveto{\pgfqpoint{0.302312in}{0.313618in}}%
\pgfpathlineto{\pgfqpoint{0.302312in}{1.113676in}}%
\pgfusepath{stroke}%
\end{pgfscope}%
\begin{pgfscope}%
\pgfsetrectcap%
\pgfsetmiterjoin%
\pgfsetlinewidth{1.254687pt}%
\definecolor{currentstroke}{rgb}{0.800000,0.800000,0.800000}%
\pgfsetstrokecolor{currentstroke}%
\pgfsetdash{}{0pt}%
\pgfpathmoveto{\pgfqpoint{1.605238in}{0.313618in}}%
\pgfpathlineto{\pgfqpoint{1.605238in}{1.113676in}}%
\pgfusepath{stroke}%
\end{pgfscope}%
\begin{pgfscope}%
\pgfsetrectcap%
\pgfsetmiterjoin%
\pgfsetlinewidth{1.254687pt}%
\definecolor{currentstroke}{rgb}{0.800000,0.800000,0.800000}%
\pgfsetstrokecolor{currentstroke}%
\pgfsetdash{}{0pt}%
\pgfpathmoveto{\pgfqpoint{0.302312in}{0.313618in}}%
\pgfpathlineto{\pgfqpoint{1.605238in}{0.313618in}}%
\pgfusepath{stroke}%
\end{pgfscope}%
\begin{pgfscope}%
\pgfsetrectcap%
\pgfsetmiterjoin%
\pgfsetlinewidth{1.254687pt}%
\definecolor{currentstroke}{rgb}{0.800000,0.800000,0.800000}%
\pgfsetstrokecolor{currentstroke}%
\pgfsetdash{}{0pt}%
\pgfpathmoveto{\pgfqpoint{0.302312in}{1.113676in}}%
\pgfpathlineto{\pgfqpoint{1.605238in}{1.113676in}}%
\pgfusepath{stroke}%
\end{pgfscope}%
\begin{pgfscope}%
\pgfsetbuttcap%
\pgfsetmiterjoin%
\definecolor{currentfill}{rgb}{1.000000,1.000000,1.000000}%
\pgfsetfillcolor{currentfill}%
\pgfsetfillopacity{0.800000}%
\pgfsetlinewidth{1.003750pt}%
\definecolor{currentstroke}{rgb}{0.800000,0.800000,0.800000}%
\pgfsetstrokecolor{currentstroke}%
\pgfsetstrokeopacity{0.800000}%
\pgfsetdash{}{0pt}%
\pgfpathmoveto{\pgfqpoint{0.350923in}{0.348341in}}%
\pgfpathlineto{\pgfqpoint{0.883254in}{0.348341in}}%
\pgfpathquadraticcurveto{\pgfqpoint{0.897143in}{0.348341in}}{\pgfqpoint{0.897143in}{0.362230in}}%
\pgfpathlineto{\pgfqpoint{0.897143in}{0.839450in}}%
\pgfpathquadraticcurveto{\pgfqpoint{0.897143in}{0.853339in}}{\pgfqpoint{0.883254in}{0.853339in}}%
\pgfpathlineto{\pgfqpoint{0.350923in}{0.853339in}}%
\pgfpathquadraticcurveto{\pgfqpoint{0.337034in}{0.853339in}}{\pgfqpoint{0.337034in}{0.839450in}}%
\pgfpathlineto{\pgfqpoint{0.337034in}{0.362230in}}%
\pgfpathquadraticcurveto{\pgfqpoint{0.337034in}{0.348341in}}{\pgfqpoint{0.350923in}{0.348341in}}%
\pgfpathlineto{\pgfqpoint{0.350923in}{0.348341in}}%
\pgfpathclose%
\pgfusepath{stroke,fill}%
\end{pgfscope}%
\begin{pgfscope}%
\pgfsetbuttcap%
\pgfsetmiterjoin%
\definecolor{currentfill}{rgb}{0.003922,0.450980,0.698039}%
\pgfsetfillcolor{currentfill}%
\pgfsetlinewidth{1.003750pt}%
\definecolor{currentstroke}{rgb}{1.000000,1.000000,1.000000}%
\pgfsetstrokecolor{currentstroke}%
\pgfsetdash{}{0pt}%
\pgfpathmoveto{\pgfqpoint{0.364812in}{0.776950in}}%
\pgfpathlineto{\pgfqpoint{0.503701in}{0.776950in}}%
\pgfpathlineto{\pgfqpoint{0.503701in}{0.825562in}}%
\pgfpathlineto{\pgfqpoint{0.364812in}{0.825562in}}%
\pgfpathlineto{\pgfqpoint{0.364812in}{0.776950in}}%
\pgfpathclose%
\pgfusepath{stroke,fill}%
\end{pgfscope}%
\begin{pgfscope}%
\definecolor{textcolor}{rgb}{0.150000,0.150000,0.150000}%
\pgfsetstrokecolor{textcolor}%
\pgfsetfillcolor{textcolor}%
\pgftext[x=0.559256in,y=0.776950in,left,base]{\color{textcolor}{\rmfamily\fontsize{5.000000}{6.000000}\selectfont\catcode`\^=\active\def^{\ifmmode\sp\else\^{}\fi}\catcode`\%=\active\def
\end{pgfscope}%
\begin{pgfscope}%
\pgfsetbuttcap%
\pgfsetmiterjoin%
\definecolor{currentfill}{rgb}{0.870588,0.560784,0.019608}%
\pgfsetfillcolor{currentfill}%
\pgfsetlinewidth{1.003750pt}%
\definecolor{currentstroke}{rgb}{1.000000,1.000000,1.000000}%
\pgfsetstrokecolor{currentstroke}%
\pgfsetdash{}{0pt}%
\pgfpathmoveto{\pgfqpoint{0.364812in}{0.680117in}}%
\pgfpathlineto{\pgfqpoint{0.503701in}{0.680117in}}%
\pgfpathlineto{\pgfqpoint{0.503701in}{0.728728in}}%
\pgfpathlineto{\pgfqpoint{0.364812in}{0.728728in}}%
\pgfpathlineto{\pgfqpoint{0.364812in}{0.680117in}}%
\pgfpathclose%
\pgfusepath{stroke,fill}%
\end{pgfscope}%
\begin{pgfscope}%
\definecolor{textcolor}{rgb}{0.150000,0.150000,0.150000}%
\pgfsetstrokecolor{textcolor}%
\pgfsetfillcolor{textcolor}%
\pgftext[x=0.559256in,y=0.680117in,left,base]{\color{textcolor}{\rmfamily\fontsize{5.000000}{6.000000}\selectfont\catcode`\^=\active\def^{\ifmmode\sp\else\^{}\fi}\catcode`\%=\active\def
\end{pgfscope}%
\begin{pgfscope}%
\pgfsetbuttcap%
\pgfsetmiterjoin%
\definecolor{currentfill}{rgb}{0.007843,0.619608,0.450980}%
\pgfsetfillcolor{currentfill}%
\pgfsetlinewidth{1.003750pt}%
\definecolor{currentstroke}{rgb}{1.000000,1.000000,1.000000}%
\pgfsetstrokecolor{currentstroke}%
\pgfsetdash{}{0pt}%
\pgfpathmoveto{\pgfqpoint{0.364812in}{0.583284in}}%
\pgfpathlineto{\pgfqpoint{0.503701in}{0.583284in}}%
\pgfpathlineto{\pgfqpoint{0.503701in}{0.631895in}}%
\pgfpathlineto{\pgfqpoint{0.364812in}{0.631895in}}%
\pgfpathlineto{\pgfqpoint{0.364812in}{0.583284in}}%
\pgfpathclose%
\pgfusepath{stroke,fill}%
\end{pgfscope}%
\begin{pgfscope}%
\definecolor{textcolor}{rgb}{0.150000,0.150000,0.150000}%
\pgfsetstrokecolor{textcolor}%
\pgfsetfillcolor{textcolor}%
\pgftext[x=0.559256in,y=0.583284in,left,base]{\color{textcolor}{\rmfamily\fontsize{5.000000}{6.000000}\selectfont\catcode`\^=\active\def^{\ifmmode\sp\else\^{}\fi}\catcode`\%=\active\def
\end{pgfscope}%
\begin{pgfscope}%
\pgfsetbuttcap%
\pgfsetmiterjoin%
\definecolor{currentfill}{rgb}{0.835294,0.368627,0.000000}%
\pgfsetfillcolor{currentfill}%
\pgfsetlinewidth{1.003750pt}%
\definecolor{currentstroke}{rgb}{1.000000,1.000000,1.000000}%
\pgfsetstrokecolor{currentstroke}%
\pgfsetdash{}{0pt}%
\pgfpathmoveto{\pgfqpoint{0.364812in}{0.486451in}}%
\pgfpathlineto{\pgfqpoint{0.503701in}{0.486451in}}%
\pgfpathlineto{\pgfqpoint{0.503701in}{0.535062in}}%
\pgfpathlineto{\pgfqpoint{0.364812in}{0.535062in}}%
\pgfpathlineto{\pgfqpoint{0.364812in}{0.486451in}}%
\pgfpathclose%
\pgfusepath{stroke,fill}%
\end{pgfscope}%
\begin{pgfscope}%
\definecolor{textcolor}{rgb}{0.150000,0.150000,0.150000}%
\pgfsetstrokecolor{textcolor}%
\pgfsetfillcolor{textcolor}%
\pgftext[x=0.559256in,y=0.486451in,left,base]{\color{textcolor}{\rmfamily\fontsize{5.000000}{6.000000}\selectfont\catcode`\^=\active\def^{\ifmmode\sp\else\^{}\fi}\catcode`\%=\active\def
\end{pgfscope}%
\begin{pgfscope}%
\pgfsetbuttcap%
\pgfsetmiterjoin%
\definecolor{currentfill}{rgb}{0.800000,0.470588,0.737255}%
\pgfsetfillcolor{currentfill}%
\pgfsetlinewidth{1.003750pt}%
\definecolor{currentstroke}{rgb}{1.000000,1.000000,1.000000}%
\pgfsetstrokecolor{currentstroke}%
\pgfsetdash{}{0pt}%
\pgfpathmoveto{\pgfqpoint{0.364812in}{0.389618in}}%
\pgfpathlineto{\pgfqpoint{0.503701in}{0.389618in}}%
\pgfpathlineto{\pgfqpoint{0.503701in}{0.438229in}}%
\pgfpathlineto{\pgfqpoint{0.364812in}{0.438229in}}%
\pgfpathlineto{\pgfqpoint{0.364812in}{0.389618in}}%
\pgfpathclose%
\pgfusepath{stroke,fill}%
\end{pgfscope}%
\begin{pgfscope}%
\definecolor{textcolor}{rgb}{0.150000,0.150000,0.150000}%
\pgfsetstrokecolor{textcolor}%
\pgfsetfillcolor{textcolor}%
\pgftext[x=0.559256in,y=0.389618in,left,base]{\color{textcolor}{\rmfamily\fontsize{5.000000}{6.000000}\selectfont\catcode`\^=\active\def^{\ifmmode\sp\else\^{}\fi}\catcode`\%=\active\def
\end{pgfscope}%
\end{pgfpicture}%
\makeatother%
\endgroup%

%% file: chapter/figures/data_statistics/Model_distribution.pgf
\begingroup%
\makeatletter%
\begin{pgfpicture}%
\pgfpathrectangle{\pgfpointorigin}{\pgfqpoint{1.655238in}{1.163676in}}%
\pgfusepath{use as bounding box, clip}%
\begin{pgfscope}%
\pgfsetbuttcap%
\pgfsetmiterjoin%
\definecolor{currentfill}{rgb}{1.000000,1.000000,1.000000}%
\pgfsetfillcolor{currentfill}%
\pgfsetlinewidth{0.000000pt}%
\definecolor{currentstroke}{rgb}{1.000000,1.000000,1.000000}%
\pgfsetstrokecolor{currentstroke}%
\pgfsetdash{}{0pt}%
\pgfpathmoveto{\pgfqpoint{0.000000in}{-0.000000in}}%
\pgfpathlineto{\pgfqpoint{1.655238in}{-0.000000in}}%
\pgfpathlineto{\pgfqpoint{1.655238in}{1.163676in}}%
\pgfpathlineto{\pgfqpoint{0.000000in}{1.163676in}}%
\pgfpathlineto{\pgfqpoint{0.000000in}{-0.000000in}}%
\pgfpathclose%
\pgfusepath{fill}%
\end{pgfscope}%
\begin{pgfscope}%
\pgfsetbuttcap%
\pgfsetmiterjoin%
\definecolor{currentfill}{rgb}{1.000000,1.000000,1.000000}%
\pgfsetfillcolor{currentfill}%
\pgfsetlinewidth{0.000000pt}%
\definecolor{currentstroke}{rgb}{0.000000,0.000000,0.000000}%
\pgfsetstrokecolor{currentstroke}%
\pgfsetstrokeopacity{0.000000}%
\pgfsetdash{}{0pt}%
\pgfpathmoveto{\pgfqpoint{0.302312in}{0.313618in}}%
\pgfpathlineto{\pgfqpoint{1.605238in}{0.313618in}}%
\pgfpathlineto{\pgfqpoint{1.605238in}{1.113676in}}%
\pgfpathlineto{\pgfqpoint{0.302312in}{1.113676in}}%
\pgfpathlineto{\pgfqpoint{0.302312in}{0.313618in}}%
\pgfpathclose%
\pgfusepath{fill}%
\end{pgfscope}%
\begin{pgfscope}%
\pgfpathrectangle{\pgfqpoint{0.302312in}{0.313618in}}{\pgfqpoint{1.302927in}{0.800058in}}%
\pgfusepath{clip}%
\pgfsetroundcap%
\pgfsetroundjoin%
\pgfsetlinewidth{1.003750pt}%
\definecolor{currentstroke}{rgb}{0.800000,0.800000,0.800000}%
\pgfsetstrokecolor{currentstroke}%
\pgfsetdash{}{0pt}%
\pgfpathmoveto{\pgfqpoint{0.509595in}{0.313618in}}%
\pgfpathlineto{\pgfqpoint{0.509595in}{1.113676in}}%
\pgfusepath{stroke}%
\end{pgfscope}%
\begin{pgfscope}%
\definecolor{textcolor}{rgb}{1.000000,1.000000,1.000000}%
\pgfsetstrokecolor{textcolor}%
\pgfsetfillcolor{textcolor}%
\pgftext[x=0.509595in,y=0.265007in,,top]{\color{textcolor}{\rmfamily\fontsize{5.000000}{6.000000}\selectfont\catcode`\^=\active\def^{\ifmmode\sp\else\^{}\fi}\catcode`\%=\active\def
\end{pgfscope}%
\begin{pgfscope}%
\pgfpathrectangle{\pgfqpoint{0.302312in}{0.313618in}}{\pgfqpoint{1.302927in}{0.800058in}}%
\pgfusepath{clip}%
\pgfsetroundcap%
\pgfsetroundjoin%
\pgfsetlinewidth{1.003750pt}%
\definecolor{currentstroke}{rgb}{0.800000,0.800000,0.800000}%
\pgfsetstrokecolor{currentstroke}%
\pgfsetdash{}{0pt}%
\pgfpathmoveto{\pgfqpoint{0.805715in}{0.313618in}}%
\pgfpathlineto{\pgfqpoint{0.805715in}{1.113676in}}%
\pgfusepath{stroke}%
\end{pgfscope}%
\begin{pgfscope}%
\definecolor{textcolor}{rgb}{1.000000,1.000000,1.000000}%
\pgfsetstrokecolor{textcolor}%
\pgfsetfillcolor{textcolor}%
\pgftext[x=0.805715in,y=0.265007in,,top]{\color{textcolor}{\rmfamily\fontsize{5.000000}{6.000000}\selectfont\catcode`\^=\active\def^{\ifmmode\sp\else\^{}\fi}\catcode`\%=\active\def
\end{pgfscope}%
\begin{pgfscope}%
\pgfpathrectangle{\pgfqpoint{0.302312in}{0.313618in}}{\pgfqpoint{1.302927in}{0.800058in}}%
\pgfusepath{clip}%
\pgfsetroundcap%
\pgfsetroundjoin%
\pgfsetlinewidth{1.003750pt}%
\definecolor{currentstroke}{rgb}{0.800000,0.800000,0.800000}%
\pgfsetstrokecolor{currentstroke}%
\pgfsetdash{}{0pt}%
\pgfpathmoveto{\pgfqpoint{1.101835in}{0.313618in}}%
\pgfpathlineto{\pgfqpoint{1.101835in}{1.113676in}}%
\pgfusepath{stroke}%
\end{pgfscope}%
\begin{pgfscope}%
\definecolor{textcolor}{rgb}{1.000000,1.000000,1.000000}%
\pgfsetstrokecolor{textcolor}%
\pgfsetfillcolor{textcolor}%
\pgftext[x=1.101835in,y=0.265007in,,top]{\color{textcolor}{\rmfamily\fontsize{5.000000}{6.000000}\selectfont\catcode`\^=\active\def^{\ifmmode\sp\else\^{}\fi}\catcode`\%=\active\def
\end{pgfscope}%
\begin{pgfscope}%
\pgfpathrectangle{\pgfqpoint{0.302312in}{0.313618in}}{\pgfqpoint{1.302927in}{0.800058in}}%
\pgfusepath{clip}%
\pgfsetroundcap%
\pgfsetroundjoin%
\pgfsetlinewidth{1.003750pt}%
\definecolor{currentstroke}{rgb}{0.800000,0.800000,0.800000}%
\pgfsetstrokecolor{currentstroke}%
\pgfsetdash{}{0pt}%
\pgfpathmoveto{\pgfqpoint{1.397954in}{0.313618in}}%
\pgfpathlineto{\pgfqpoint{1.397954in}{1.113676in}}%
\pgfusepath{stroke}%
\end{pgfscope}%
\begin{pgfscope}%
\definecolor{textcolor}{rgb}{1.000000,1.000000,1.000000}%
\pgfsetstrokecolor{textcolor}%
\pgfsetfillcolor{textcolor}%
\pgftext[x=1.397954in,y=0.265007in,,top]{\color{textcolor}{\rmfamily\fontsize{5.000000}{6.000000}\selectfont\catcode`\^=\active\def^{\ifmmode\sp\else\^{}\fi}\catcode`\%=\active\def
\end{pgfscope}%
\begin{pgfscope}%
\definecolor{textcolor}{rgb}{0.150000,0.150000,0.150000}%
\pgfsetstrokecolor{textcolor}%
\pgfsetfillcolor{textcolor}%
\pgftext[x=0.953775in,y=0.148124in,,top]{\color{textcolor}{\rmfamily\fontsize{8.000000}{9.600000}\selectfont\catcode`\^=\active\def^{\ifmmode\sp\else\^{}\fi}\catcode`\%=\active\def
\end{pgfscope}%
\begin{pgfscope}%
\pgfpathrectangle{\pgfqpoint{0.302312in}{0.313618in}}{\pgfqpoint{1.302927in}{0.800058in}}%
\pgfusepath{clip}%
\pgfsetroundcap%
\pgfsetroundjoin%
\pgfsetlinewidth{1.003750pt}%
\definecolor{currentstroke}{rgb}{0.800000,0.800000,0.800000}%
\pgfsetstrokecolor{currentstroke}%
\pgfsetdash{}{0pt}%
\pgfpathmoveto{\pgfqpoint{0.302312in}{0.313618in}}%
\pgfpathlineto{\pgfqpoint{1.605238in}{0.313618in}}%
\pgfusepath{stroke}%
\end{pgfscope}%
\begin{pgfscope}%
\definecolor{textcolor}{rgb}{0.150000,0.150000,0.150000}%
\pgfsetstrokecolor{textcolor}%
\pgfsetfillcolor{textcolor}%
\pgftext[x=0.202775in, y=0.284690in, left, base]{\color{textcolor}{\rmfamily\fontsize{6.000000}{7.200000}\selectfont\catcode`\^=\active\def^{\ifmmode\sp\else\^{}\fi}\catcode`\%=\active\def
\end{pgfscope}%
\begin{pgfscope}%
\pgfpathrectangle{\pgfqpoint{0.302312in}{0.313618in}}{\pgfqpoint{1.302927in}{0.800058in}}%
\pgfusepath{clip}%
\pgfsetroundcap%
\pgfsetroundjoin%
\pgfsetlinewidth{1.003750pt}%
\definecolor{currentstroke}{rgb}{0.800000,0.800000,0.800000}%
\pgfsetstrokecolor{currentstroke}%
\pgfsetdash{}{0pt}%
\pgfpathmoveto{\pgfqpoint{0.302312in}{0.569095in}}%
\pgfpathlineto{\pgfqpoint{1.605238in}{0.569095in}}%
\pgfusepath{stroke}%
\end{pgfscope}%
\begin{pgfscope}%
\definecolor{textcolor}{rgb}{0.150000,0.150000,0.150000}%
\pgfsetstrokecolor{textcolor}%
\pgfsetfillcolor{textcolor}%
\pgftext[x=0.100925in, y=0.540167in, left, base]{\color{textcolor}{\rmfamily\fontsize{6.000000}{7.200000}\selectfont\catcode`\^=\active\def^{\ifmmode\sp\else\^{}\fi}\catcode`\%=\active\def
\end{pgfscope}%
\begin{pgfscope}%
\pgfpathrectangle{\pgfqpoint{0.302312in}{0.313618in}}{\pgfqpoint{1.302927in}{0.800058in}}%
\pgfusepath{clip}%
\pgfsetroundcap%
\pgfsetroundjoin%
\pgfsetlinewidth{1.003750pt}%
\definecolor{currentstroke}{rgb}{0.800000,0.800000,0.800000}%
\pgfsetstrokecolor{currentstroke}%
\pgfsetdash{}{0pt}%
\pgfpathmoveto{\pgfqpoint{0.302312in}{0.824572in}}%
\pgfpathlineto{\pgfqpoint{1.605238in}{0.824572in}}%
\pgfusepath{stroke}%
\end{pgfscope}%
\begin{pgfscope}%
\definecolor{textcolor}{rgb}{0.150000,0.150000,0.150000}%
\pgfsetstrokecolor{textcolor}%
\pgfsetfillcolor{textcolor}%
\pgftext[x=0.050000in, y=0.795644in, left, base]{\color{textcolor}{\rmfamily\fontsize{6.000000}{7.200000}\selectfont\catcode`\^=\active\def^{\ifmmode\sp\else\^{}\fi}\catcode`\%=\active\def
\end{pgfscope}%
\begin{pgfscope}%
\pgfpathrectangle{\pgfqpoint{0.302312in}{0.313618in}}{\pgfqpoint{1.302927in}{0.800058in}}%
\pgfusepath{clip}%
\pgfsetroundcap%
\pgfsetroundjoin%
\pgfsetlinewidth{1.003750pt}%
\definecolor{currentstroke}{rgb}{0.800000,0.800000,0.800000}%
\pgfsetstrokecolor{currentstroke}%
\pgfsetdash{}{0pt}%
\pgfpathmoveto{\pgfqpoint{0.302312in}{1.080049in}}%
\pgfpathlineto{\pgfqpoint{1.605238in}{1.080049in}}%
\pgfusepath{stroke}%
\end{pgfscope}%
\begin{pgfscope}%
\definecolor{textcolor}{rgb}{0.150000,0.150000,0.150000}%
\pgfsetstrokecolor{textcolor}%
\pgfsetfillcolor{textcolor}%
\pgftext[x=0.050000in, y=1.051121in, left, base]{\color{textcolor}{\rmfamily\fontsize{6.000000}{7.200000}\selectfont\catcode`\^=\active\def^{\ifmmode\sp\else\^{}\fi}\catcode`\%=\active\def
\end{pgfscope}%
\begin{pgfscope}%
\pgfpathrectangle{\pgfqpoint{0.302312in}{0.313618in}}{\pgfqpoint{1.302927in}{0.800058in}}%
\pgfusepath{clip}%
\pgfsetbuttcap%
\pgfsetmiterjoin%
\definecolor{currentfill}{rgb}{0.003922,0.450980,0.698039}%
\pgfsetfillcolor{currentfill}%
\pgfsetlinewidth{0.501875pt}%
\definecolor{currentstroke}{rgb}{1.000000,1.000000,1.000000}%
\pgfsetstrokecolor{currentstroke}%
\pgfsetdash{}{0pt}%
\pgfpathmoveto{\pgfqpoint{0.361536in}{0.313618in}}%
\pgfpathlineto{\pgfqpoint{0.657655in}{0.313618in}}%
\pgfpathlineto{\pgfqpoint{0.657655in}{1.075578in}}%
\pgfpathlineto{\pgfqpoint{0.361536in}{1.075578in}}%
\pgfpathlineto{\pgfqpoint{0.361536in}{0.313618in}}%
\pgfpathclose%
\pgfusepath{stroke,fill}%
\end{pgfscope}%
\begin{pgfscope}%
\pgfpathrectangle{\pgfqpoint{0.302312in}{0.313618in}}{\pgfqpoint{1.302927in}{0.800058in}}%
\pgfusepath{clip}%
\pgfsetbuttcap%
\pgfsetmiterjoin%
\definecolor{currentfill}{rgb}{0.870588,0.560784,0.019608}%
\pgfsetfillcolor{currentfill}%
\pgfsetlinewidth{0.501875pt}%
\definecolor{currentstroke}{rgb}{1.000000,1.000000,1.000000}%
\pgfsetstrokecolor{currentstroke}%
\pgfsetdash{}{0pt}%
\pgfpathmoveto{\pgfqpoint{0.657655in}{0.313618in}}%
\pgfpathlineto{\pgfqpoint{0.953775in}{0.313618in}}%
\pgfpathlineto{\pgfqpoint{0.953775in}{0.921015in}}%
\pgfpathlineto{\pgfqpoint{0.657655in}{0.921015in}}%
\pgfpathlineto{\pgfqpoint{0.657655in}{0.313618in}}%
\pgfpathclose%
\pgfusepath{stroke,fill}%
\end{pgfscope}%
\begin{pgfscope}%
\pgfpathrectangle{\pgfqpoint{0.302312in}{0.313618in}}{\pgfqpoint{1.302927in}{0.800058in}}%
\pgfusepath{clip}%
\pgfsetbuttcap%
\pgfsetmiterjoin%
\definecolor{currentfill}{rgb}{0.007843,0.619608,0.450980}%
\pgfsetfillcolor{currentfill}%
\pgfsetlinewidth{0.501875pt}%
\definecolor{currentstroke}{rgb}{1.000000,1.000000,1.000000}%
\pgfsetstrokecolor{currentstroke}%
\pgfsetdash{}{0pt}%
\pgfpathmoveto{\pgfqpoint{0.953775in}{0.313618in}}%
\pgfpathlineto{\pgfqpoint{1.249895in}{0.313618in}}%
\pgfpathlineto{\pgfqpoint{1.249895in}{0.814672in}}%
\pgfpathlineto{\pgfqpoint{0.953775in}{0.814672in}}%
\pgfpathlineto{\pgfqpoint{0.953775in}{0.313618in}}%
\pgfpathclose%
\pgfusepath{stroke,fill}%
\end{pgfscope}%
\begin{pgfscope}%
\pgfpathrectangle{\pgfqpoint{0.302312in}{0.313618in}}{\pgfqpoint{1.302927in}{0.800058in}}%
\pgfusepath{clip}%
\pgfsetbuttcap%
\pgfsetmiterjoin%
\definecolor{currentfill}{rgb}{0.835294,0.368627,0.000000}%
\pgfsetfillcolor{currentfill}%
\pgfsetlinewidth{0.501875pt}%
\definecolor{currentstroke}{rgb}{1.000000,1.000000,1.000000}%
\pgfsetstrokecolor{currentstroke}%
\pgfsetdash{}{0pt}%
\pgfpathmoveto{\pgfqpoint{1.249895in}{0.313618in}}%
\pgfpathlineto{\pgfqpoint{1.546014in}{0.313618in}}%
\pgfpathlineto{\pgfqpoint{1.546014in}{0.564305in}}%
\pgfpathlineto{\pgfqpoint{1.249895in}{0.564305in}}%
\pgfpathlineto{\pgfqpoint{1.249895in}{0.313618in}}%
\pgfpathclose%
\pgfusepath{stroke,fill}%
\end{pgfscope}%
\begin{pgfscope}%
\pgfsetrectcap%
\pgfsetmiterjoin%
\pgfsetlinewidth{1.254687pt}%
\definecolor{currentstroke}{rgb}{0.800000,0.800000,0.800000}%
\pgfsetstrokecolor{currentstroke}%
\pgfsetdash{}{0pt}%
\pgfpathmoveto{\pgfqpoint{0.302312in}{0.313618in}}%
\pgfpathlineto{\pgfqpoint{0.302312in}{1.113676in}}%
\pgfusepath{stroke}%
\end{pgfscope}%
\begin{pgfscope}%
\pgfsetrectcap%
\pgfsetmiterjoin%
\pgfsetlinewidth{1.254687pt}%
\definecolor{currentstroke}{rgb}{0.800000,0.800000,0.800000}%
\pgfsetstrokecolor{currentstroke}%
\pgfsetdash{}{0pt}%
\pgfpathmoveto{\pgfqpoint{1.605238in}{0.313618in}}%
\pgfpathlineto{\pgfqpoint{1.605238in}{1.113676in}}%
\pgfusepath{stroke}%
\end{pgfscope}%
\begin{pgfscope}%
\pgfsetrectcap%
\pgfsetmiterjoin%
\pgfsetlinewidth{1.254687pt}%
\definecolor{currentstroke}{rgb}{0.800000,0.800000,0.800000}%
\pgfsetstrokecolor{currentstroke}%
\pgfsetdash{}{0pt}%
\pgfpathmoveto{\pgfqpoint{0.302312in}{0.313618in}}%
\pgfpathlineto{\pgfqpoint{1.605238in}{0.313618in}}%
\pgfusepath{stroke}%
\end{pgfscope}%
\begin{pgfscope}%
\pgfsetrectcap%
\pgfsetmiterjoin%
\pgfsetlinewidth{1.254687pt}%
\definecolor{currentstroke}{rgb}{0.800000,0.800000,0.800000}%
\pgfsetstrokecolor{currentstroke}%
\pgfsetdash{}{0pt}%
\pgfpathmoveto{\pgfqpoint{0.302312in}{1.113676in}}%
\pgfpathlineto{\pgfqpoint{1.605238in}{1.113676in}}%
\pgfusepath{stroke}%
\end{pgfscope}%
\begin{pgfscope}%
\pgfsetbuttcap%
\pgfsetmiterjoin%
\definecolor{currentfill}{rgb}{1.000000,1.000000,1.000000}%
\pgfsetfillcolor{currentfill}%
\pgfsetfillopacity{0.800000}%
\pgfsetlinewidth{1.003750pt}%
\definecolor{currentstroke}{rgb}{0.800000,0.800000,0.800000}%
\pgfsetstrokecolor{currentstroke}%
\pgfsetstrokeopacity{0.800000}%
\pgfsetdash{}{0pt}%
\pgfpathmoveto{\pgfqpoint{0.350923in}{0.348341in}}%
\pgfpathlineto{\pgfqpoint{1.079098in}{0.348341in}}%
\pgfpathquadraticcurveto{\pgfqpoint{1.092987in}{0.348341in}}{\pgfqpoint{1.092987in}{0.362230in}}%
\pgfpathlineto{\pgfqpoint{1.092987in}{0.742617in}}%
\pgfpathquadraticcurveto{\pgfqpoint{1.092987in}{0.756506in}}{\pgfqpoint{1.079098in}{0.756506in}}%
\pgfpathlineto{\pgfqpoint{0.350923in}{0.756506in}}%
\pgfpathquadraticcurveto{\pgfqpoint{0.337034in}{0.756506in}}{\pgfqpoint{0.337034in}{0.742617in}}%
\pgfpathlineto{\pgfqpoint{0.337034in}{0.362230in}}%
\pgfpathquadraticcurveto{\pgfqpoint{0.337034in}{0.348341in}}{\pgfqpoint{0.350923in}{0.348341in}}%
\pgfpathlineto{\pgfqpoint{0.350923in}{0.348341in}}%
\pgfpathclose%
\pgfusepath{stroke,fill}%
\end{pgfscope}%
\begin{pgfscope}%
\pgfsetbuttcap%
\pgfsetmiterjoin%
\definecolor{currentfill}{rgb}{0.003922,0.450980,0.698039}%
\pgfsetfillcolor{currentfill}%
\pgfsetlinewidth{1.003750pt}%
\definecolor{currentstroke}{rgb}{1.000000,1.000000,1.000000}%
\pgfsetstrokecolor{currentstroke}%
\pgfsetdash{}{0pt}%
\pgfpathmoveto{\pgfqpoint{0.364812in}{0.680117in}}%
\pgfpathlineto{\pgfqpoint{0.503701in}{0.680117in}}%
\pgfpathlineto{\pgfqpoint{0.503701in}{0.728728in}}%
\pgfpathlineto{\pgfqpoint{0.364812in}{0.728728in}}%
\pgfpathlineto{\pgfqpoint{0.364812in}{0.680117in}}%
\pgfpathclose%
\pgfusepath{stroke,fill}%
\end{pgfscope}%
\begin{pgfscope}%
\definecolor{textcolor}{rgb}{0.150000,0.150000,0.150000}%
\pgfsetstrokecolor{textcolor}%
\pgfsetfillcolor{textcolor}%
\pgftext[x=0.559256in,y=0.680117in,left,base]{\color{textcolor}{\rmfamily\fontsize{5.000000}{6.000000}\selectfont\catcode`\^=\active\def^{\ifmmode\sp\else\^{}\fi}\catcode`\%=\active\def
\end{pgfscope}%
\begin{pgfscope}%
\pgfsetbuttcap%
\pgfsetmiterjoin%
\definecolor{currentfill}{rgb}{0.870588,0.560784,0.019608}%
\pgfsetfillcolor{currentfill}%
\pgfsetlinewidth{1.003750pt}%
\definecolor{currentstroke}{rgb}{1.000000,1.000000,1.000000}%
\pgfsetstrokecolor{currentstroke}%
\pgfsetdash{}{0pt}%
\pgfpathmoveto{\pgfqpoint{0.364812in}{0.583284in}}%
\pgfpathlineto{\pgfqpoint{0.503701in}{0.583284in}}%
\pgfpathlineto{\pgfqpoint{0.503701in}{0.631895in}}%
\pgfpathlineto{\pgfqpoint{0.364812in}{0.631895in}}%
\pgfpathlineto{\pgfqpoint{0.364812in}{0.583284in}}%
\pgfpathclose%
\pgfusepath{stroke,fill}%
\end{pgfscope}%
\begin{pgfscope}%
\definecolor{textcolor}{rgb}{0.150000,0.150000,0.150000}%
\pgfsetstrokecolor{textcolor}%
\pgfsetfillcolor{textcolor}%
\pgftext[x=0.559256in,y=0.583284in,left,base]{\color{textcolor}{\rmfamily\fontsize{5.000000}{6.000000}\selectfont\catcode`\^=\active\def^{\ifmmode\sp\else\^{}\fi}\catcode`\%=\active\def
\end{pgfscope}%
\begin{pgfscope}%
\pgfsetbuttcap%
\pgfsetmiterjoin%
\definecolor{currentfill}{rgb}{0.007843,0.619608,0.450980}%
\pgfsetfillcolor{currentfill}%
\pgfsetlinewidth{1.003750pt}%
\definecolor{currentstroke}{rgb}{1.000000,1.000000,1.000000}%
\pgfsetstrokecolor{currentstroke}%
\pgfsetdash{}{0pt}%
\pgfpathmoveto{\pgfqpoint{0.364812in}{0.486451in}}%
\pgfpathlineto{\pgfqpoint{0.503701in}{0.486451in}}%
\pgfpathlineto{\pgfqpoint{0.503701in}{0.535062in}}%
\pgfpathlineto{\pgfqpoint{0.364812in}{0.535062in}}%
\pgfpathlineto{\pgfqpoint{0.364812in}{0.486451in}}%
\pgfpathclose%
\pgfusepath{stroke,fill}%
\end{pgfscope}%
\begin{pgfscope}%
\definecolor{textcolor}{rgb}{0.150000,0.150000,0.150000}%
\pgfsetstrokecolor{textcolor}%
\pgfsetfillcolor{textcolor}%
\pgftext[x=0.559256in,y=0.486451in,left,base]{\color{textcolor}{\rmfamily\fontsize{5.000000}{6.000000}\selectfont\catcode`\^=\active\def^{\ifmmode\sp\else\^{}\fi}\catcode`\%=\active\def
\end{pgfscope}%
\begin{pgfscope}%
\pgfsetbuttcap%
\pgfsetmiterjoin%
\definecolor{currentfill}{rgb}{0.835294,0.368627,0.000000}%
\pgfsetfillcolor{currentfill}%
\pgfsetlinewidth{1.003750pt}%
\definecolor{currentstroke}{rgb}{1.000000,1.000000,1.000000}%
\pgfsetstrokecolor{currentstroke}%
\pgfsetdash{}{0pt}%
\pgfpathmoveto{\pgfqpoint{0.364812in}{0.389618in}}%
\pgfpathlineto{\pgfqpoint{0.503701in}{0.389618in}}%
\pgfpathlineto{\pgfqpoint{0.503701in}{0.438229in}}%
\pgfpathlineto{\pgfqpoint{0.364812in}{0.438229in}}%
\pgfpathlineto{\pgfqpoint{0.364812in}{0.389618in}}%
\pgfpathclose%
\pgfusepath{stroke,fill}%
\end{pgfscope}%
\begin{pgfscope}%
\definecolor{textcolor}{rgb}{0.150000,0.150000,0.150000}%
\pgfsetstrokecolor{textcolor}%
\pgfsetfillcolor{textcolor}%
\pgftext[x=0.559256in,y=0.389618in,left,base]{\color{textcolor}{\rmfamily\fontsize{5.000000}{6.000000}\selectfont\catcode`\^=\active\def^{\ifmmode\sp\else\^{}\fi}\catcode`\%=\active\def
\end{pgfscope}%
\end{pgfpicture}%
\makeatother%
\endgroup%

%% file: chapter/4_results.tex
\section{Results}

Here, we investigate whether LLM judges maintain reliability when subjected to the distribution shifts inherent in adversarial robustness evaluations.


\begin{figure}[!t]
    \centering
    \subfigure[Attack]{
        \includegraphics[width=0.45\linewidth]{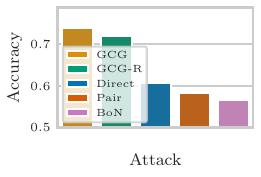}
    }
    \subfigure[Model]{
        \includegraphics[width=0.45\linewidth]{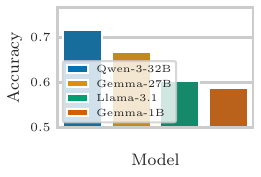}
    }
    \subfigure[Behavior]{
        \includegraphics[width=0.45\linewidth]{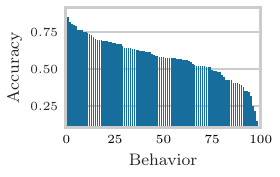}
    }
    \subfigure[Category]{
        \includegraphics[width=0.45\linewidth]{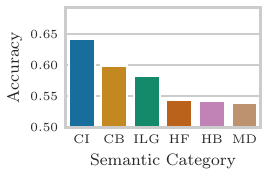}
    }
    \caption{Average judge accuracy across different distribution shifts: (a) Attack, (b) Model, (c) Behavior, (d) Semantic Category.}
    \label{fig:bar_accuracy}
    \vspace{-0.35cm}
\end{figure}

\subsection{Judge Failures}\label{sec:judge_failures}

We systematically evaluate judge performance under three critical distribution shifts. \textbf{Attack Shift:} adversarial prompts may produce distorted, high-perplexity outputs that differ from the harmful responses judges are trained on. \textbf{Model Shift}: judges validated on one model's outputs degrade when applied to different models and defenses due to linguistic variations. \textbf{Data Shift}: the difficulty of judging varies by category; detecting subtle propaganda may be harder than detecting explicit violence.


\textbf{Attack and Model Shift.} For the first experiment, we balance our dataset across every model-attack pair to ensure an equal distribution of harmful and harmless examples (based on human ground-truth labels). This controlled setup results in a dataset of $2746$ samples. Figure~\ref{fig:heatmap} visualizes judge accuracy across various attacks for \llamaeightb{} and \gemmalarge{} (see Appendix~\ref{app:extended_results} for \gemmaoneb{} and \qwen{}). Notably, performance is highly victim model dependent, with many judges performing close to random guessing. This confirms that high validation scores on benign or standard-harmful data can fail to generalize to the distribution shifts inherent in adversarial evaluations.

Furthermore, accuracy is heavily influenced by the specific attack-victim-model combination. For example, while Direct Prompting is among the hardest to judge for outputs from \llamaeightb{}, with accuracies as low as $45\%$ using the HarmBench judge, they prove considerably easier for outputs from \gemmalarge{}, reaching accuracies as high as $58\%$ with the same judge.
This demonstrates that the effect of an attack shift is not universal; rather, it emerges from the unique interaction between the attack strategy and the specific response characteristics of the victim model.

\textbf{Aggregated Results \& Data Shifts.} Figure~\ref{fig:bar_accuracy} provides an aggregated view of these performance shifts averaged across judges and other dimensions. Subfigure~(a) demonstrates considerable differences in \textit{judgeability} between attack types. Similarly, subfigure~(b) shows that the judgeability of the victim models diverges with the outputs of smaller models being slightly more difficult to assess than larger models. Subfigure~(c) shows that individual behaviors exhibit significantly different judge accuracy, suggesting that certain behaviors may be unsuitable for inclusion in reliable academic benchmarks. Moreover, subfigure~(d) shows that judge accuracy varies across semantic categories (see Appendix~\ref{app:semantic} for description). This discrepancy suggests that current safety evaluations may be implicitly biased toward detecting harms in categories that are easier for judges to classify, potentially overlooking more subtle or high-risk threats that fall outside these distributions.

\begin{figure}[t!]
    \centering
    \subfigure{
        \includegraphics[width=0.95\linewidth]{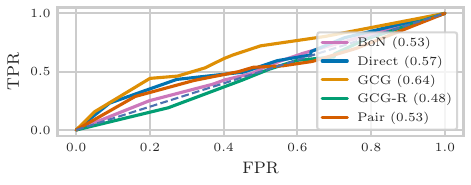}
    }
    \caption{ROC curves and AUROC scores for JailJudge on generations from the \llamaeightb{} model.}
    \label{fig:roc_curves}
\end{figure}

\textbf{Threshold Independence.} Finally, we investigate whether the observed judge failures stem from suboptimal decision thresholds or a fundamental inability to distinguish between classes. Figure~\ref{fig:roc_curves} displays the Receiver Operating Characteristic (ROC) curves for JailJudge across various attack distributions and the \llamaeightb{} model (see Appendix~\ref{ROC Curves} for more judges and models). We find that adjusting the classification threshold is insufficient to resolve effects from distribution shifts; the curves largely hover around the diagonal, indicating performance near random chance. Further, the judges consistently achieve low Area Under the ROC Curve (AUROC) scores across different attacks. JailJudge scores reach a minimum of 0.48 (GCG-R), which is worse than random guessing, and a maximum of 0.64 (GCG), failing to meet the reliability reported in prior work.

\textbf{Correlation with Human Ratings.} While we binarized human labels into \enquote{harmful} and \enquote{benign} categories in the previous experiments, the following analysis examines the granular correlation between automated judge scores and human ratings across the full evaluative scale.

\begin{figure}[!h]
    \centering
    \subfigure[HarmBench]{
        \includegraphics[width=0.45\linewidth]{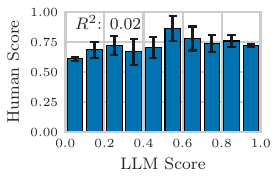}
    }
    \subfigure[Average]{
        \includegraphics[width=0.45\linewidth]{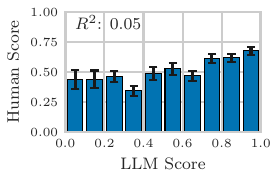}
    }
    \caption{Correlation between human ratings and judge scores for \llamaeightb{}. (a) Individual judge, (b) Averaged judge scores.}
    \label{fig:correlation_bar}
    \vspace{-0.cm}
\end{figure}

Figure~\ref{fig:correlation_bar} compares the correlation of normalized judge scores against human ratings for \llamaeightb{} (see Appendix~\ref{app:correlation} for all models). We do not observe a consistent connection between individual judge scores and human assessments (Figure~\ref{fig:correlation_bar}a), indicating that single judges are noisy. Furthermore, even averaging results over multiple judges (Figure~\ref{fig:correlation_bar}b) does not yield a clear correlation, suggesting that ensemble methods are insufficient to mitigate the systematic failures of individual judges. We further investigate ensembles in Section~\ref{sec:judge_agreement}.

\begin{tcolorbox}[
    colback=white, colframe=customblue, coltitle=white, fonttitle=\bfseries, 
    rounded corners, enhanced, 
    title=Takeaway, 
    attach boxed title to top left={yshift=-2mm, xshift=5mm}, 
    boxed title style={colback=customblue, rounded corners},
    boxsep=1mm,
    left=1mm,
    right=1mm
]
In adversarial settings, judges perform close to random chance, considerably worse than prior work suggests.
\end{tcolorbox}

\subsection{Influence of Attack Optimization}\label{sec:attack_opt}
We analyze the dynamics of judge behavior throughout the attack optimization process to determine whether adversarial strategies genuinely increase human-rated harmfulness or merely exploit insufficiencies of judges. An intuitive assumption might be that as an attack is optimized, it should produce outputs that are more \enquote{clearly} harmful and, consequently, easier for a judge to classify.

\begin{figure*}
    \centering
    \subfigure[Actual Harmfulness]{
        \includegraphics[width=0.31\linewidth]{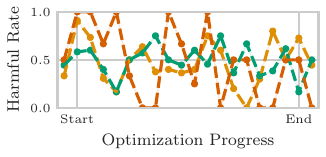}
    }
    \subfigure[Judged Harmfulness]{
        \includegraphics[width=0.31\linewidth]{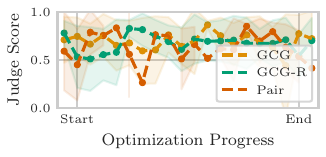}
    }
    \subfigure[Judge Difficulty]{
        \includegraphics[width=0.31\linewidth]{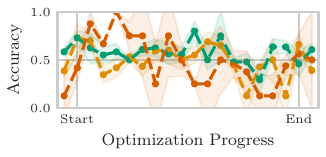}
    }
    \caption{Different metrics tracked over the optimization trajectory of several attacks.}
    \label{fig:correlation_steps}
\end{figure*}

Yet, Figure~\ref{fig:correlation_steps} contradicts this intuition. We observe no clear connection between the optimization stage and human-rated harmfulness (Figure~\ref{fig:correlation_steps}a), which remains largely stagnant. Furthermore, we do not observe a consistent increase in judge scores (Figure~\ref{fig:correlation_steps}b), nor do the attacks become easier to classify as they progress and judge accuracy fluctuates or remains flat (Figure~\ref{fig:correlation_steps}c). These results suggest that current optimization strategies may be targeting judge-specific noise or reward signals that lack a meaningful correlation with the true adversarial objective. We provide further discussion, including the relatively high harmfulness rate at the beginning of attack optimization in subfigure a) in Appendix~\ref{app:attackopt}.

\begin{tcolorbox}[
    colback=white, colframe=customblue, coltitle=white, fonttitle=\bfseries, 
    rounded corners, enhanced, 
    title=Takeaway, 
    attach boxed title to top left={yshift=-2mm, xshift=5mm}, 
    boxed title style={colback=customblue, rounded corners},
    boxsep=1mm,
    left=1mm,
    right=1mm
]
Adversarial attacks do not consistently make outputs more harmful, nor do outputs get easier to judge at late optimization stages.
\end{tcolorbox}

\subsection{Influence on Attack Success}\label{sec:attack_success}

\begin{figure}
    \centering
    \subfigure[Expected True Positives]{
        \includegraphics[width=0.47\linewidth]{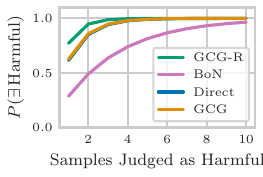}
    }
    \subfigure[Corrected ASR]{
        \includegraphics[width=0.47\linewidth]{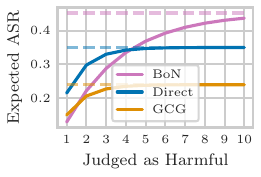}
    }
    \caption{Impact of judge errors on attack success evaluation measured on the \gemmaoneb{} model. (a) The probability of finding at least one true positive given $N$ judge-positive samples. (b) Comparison of uncorrected ASR (as reported in prior work) versus ASR corrected for judge precision. The gap illustrates how judge failures inflate perceived attack success.}
    \label{fig:attack_success}
    \vspace{-0.5cm}
\end{figure}

The errors in judge ratings identified above have considerable implications for evaluating adversarial robustness. Standard evaluations consider an attack successful if \textit{any} generated sample is flagged by the judge. Figure~\ref{fig:attack_success}(a) challenges this practice by showing the probability that at least one sample is a True Positive given $N$ judge-positive samples, denoted $P(\exists \text{Harmful})$. For attacks like BoN, a single judge-positive sample provides very low confidence that the model has actually been broken, and considerably more samples are required to guarantee at least one successful jailbreak (see Appendix~\ref{app:expected} for results across models). 

Figure~\ref{fig:attack_success}(b) quantifies the impact of these errors on the reported Attack Success Rate (ASR). When we correct for judge precision—adjusting the success rate by the probability that a judge-positive is actually a true positive, the \textit{Expected ASR} drops significantly for several methods compared to the uncorrected ASR values (shown as the horizontal dashed lines). This demonstrates that reported improvements in attacks or defenses in the literature may often be artifacts of judge failure modes (e.g., exploiting false positives) rather than genuine progress. In Figure~\ref{fig:best_to_worst}, we show that the attack, which would have been reported as the best in this specific case (BoN), achieves the lowest ASR when we correct for false positives by multiplying the ASR by judge precision. \textbf{Crucially}, our experiments demonstrate an actionable solution: evaluations become more reliable when multiple judge-positive samples are collected per behavior rather than stopping at the first positive judgment.

\begin{tcolorbox}[
    colback=white, colframe=customblue, coltitle=white, fonttitle=\bfseries, 
    rounded corners, enhanced, 
    title=Takeaway, 
    attach boxed title to top left={yshift=-2mm, xshift=5mm}, 
    boxed title style={colback=customblue, rounded corners},
    boxsep=1mm,
    left=1mm,
    right=1mm
]
Judge failures can inflate ASR. Once corrected for precision, many attacks prove far less effective.
\end{tcolorbox}

\subsection{Measuring Inter-judge Agreement}\label{sec:judge_agreement}

In this section, we introduce an inter-judge agreement signal, the Judge Concordance score, that summarizes the consistency of a panel of safety judges. While prior work relies on pairwise statistics (e.g., Cohen’s Kappa~\citep{shen2025pandaguard0}), this score is designed to capture multi-judge consensus in a single, interpretable value. We then evaluate whether high inter-judge agreement correlates positively with human-grounded correctness. 

\textbf{Judge Concordance score.} We define a score that measures the degree of consensus among 
$N$ judges for a given (prompt, generation) pair $x$.
    Let \mbox{$y_x^i \in \{0,1\}$} denote the decision of judge $i$ on $x$, where $y_x^i = 1$ if judge $i$ labels $x$ as harmful and $y_x^i = 0$ otherwise. Let $N$ be the total number of judges, and define $
        p_x = \frac{1}{N} \sum_{i=1}^N y_x^i $
    as the empirical fraction of judges that classify $x$ as harmful. The binary entropy of this distribution is given by:
    \[
        H(p_x) = -\big(p_x \log_2 p_x + (1 - p_x)\log_2(1 - p_x)\big).
    \] We then define the Judge Concordance score as the complement of the entropy:
    \[
    S_{JC}(x) = 1 - H(p_x).
    \]
Under this formulation, $S_{JC}(x) \in [0,1]$, where a score of $1$ indicates unanimous agreement, while a score of $0$ indicates maximum ambiguity in judgements. Note that since we use the logarithm base 2 for binary entropy, the score is naturally normalized between 0 and 1.

\paragraph{Does agreement predict correctness?} 
For each behavior, we compute the mean Judge Concordance score and compare that against the average accuracy of judges for instances in that behavior. Our analysis shows that \textbf{agreement does not reliably imply correctness} (see also Appendix~\ref{app:concordance}, Figure~\ref{fig:app_judge_concordance_score_heatmap}). 
Notably, across several behaviors, judges achieve near-unanimous consensus (mean concordance score close to $1$) yet consistently fail to align with human ground truth. This phenomenon suggests that safety judges may share systematic failure modes, rendering simple consensus an insufficient metric for evaluating model reliability. 

While we found that concordance does not serve as a direct proxy for correctness, the Judge Concordance score still offers a quantifiable metric for judge reliability that does not depend on human labeling. It could be used as a valuable diagnostic tool to identify behaviors for which judges share common logic or to flag ambiguous instances that may warrant targeted investigation. 

\begin{tcolorbox}[
    colback=white, colframe=customblue, coltitle=white, fonttitle=\bfseries, 
    rounded corners, enhanced, 
    title=Takeaway, 
    attach boxed title to top left={yshift=-2mm, xshift=5mm}, 
    boxed title style={colback=customblue, rounded corners},
    boxsep=1mm,
    left=1mm,
    right=1mm
]
Judge ensembles are insufficient to mitigate failures.
\end{tcolorbox}

\subsection{New Benchmarks}\label{sec:new_benchmarks}

To address the reliability issues identified in our analysis, we introduce two new datasets designed to improve future safety evaluations: \textbf{ReliableBench (RB)}, an evaluation set comprising instances with strong cross-judge agreement, and \textbf{JudgeStressTest (JST)}, a challenging set that induces substantial disagreement among judge models.

\begin{figure}[t]
    \centering
    \subfigure{
        \includegraphics[width=0.95\linewidth]{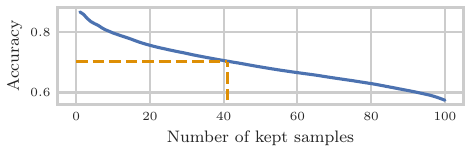}
    }
    \caption{Analysis of judge reliability for the construction of ReliableBench. (a) The average judge accuracy decreases as more behaviors are added to the dataset (sorted by difficulty). Retaining the top 41 behaviors improves accuracy to $70\%$ from $53\%$.}
    \label{fig:reliable_bench}
\end{figure}

\textbf{ReliableBench.}
To construct a more reliable evaluation subset, we analyze the relationship between the number of included behaviors and the resulting judge accuracy averaged over all judges. Figure~\ref{fig:reliable_bench} illustrates the evolution of judge accuracy as we progressively add behaviors to the dataset, sorted by their average judge accuracy (from easiest to hardest). We observe a monotonic decrease in accuracy as more difficult behaviors are included. Notably, by selecting the subset of the 41 most consistent behaviors, we can increase the average judge accuracy from $57\%$ to $70\%$.

We further analyze the distribution of judge accuracy across different models and attacks to verify how robust these reliable behaviors generalize (see also Figure~\ref{fig:reliable_bench_appendix} in Appendix~\ref{app:reliablebench}). While there is a spread in accuracy due to the variability introduced by different attacks and models, the overall mean accuracy gradually decreases with relatively low standard deviation, as more behaviors are added, indicating that the \textit{judgeability} of ReliableBench generalizes. 

\textbf{JudgeStressTest.} To investigate judge robustness on edge cases, we first construct a balanced dataset across harmful and benign human labels for every attack-model combination. We then isolate a challenging subset containing only samples for which at most one judge correctly predicted the human label. To determine if this filtering identifies samples with \textit{generalizable difficulty}, we use a leave-one-out cross-validation strategy. For each judge, we isolate the Stress Test subset based solely on the remaining performance of the remaining judges. We then measure the accuracy of the held-out judge on instances where either zero or exactly one of the other judges matched the human label.

The results, summarized in Table~\ref{tab:judge_stress_test}, indicate that difficulty is highly transferable across architectures. When all other judges fail (\texttt{Acc@0}), the held-out judge's accuracy decreases considerably relative to its overall performance on the full dataset. This suggests that JST successfully isolates systemic failure cases rather than model-specific errors. The final benchmark contains $971$ samples.

\begin{table}[ht]
\vspace{-0.5cm}
\centering
\vskip 0.15in
\begin{small}
\begin{sc}
\begin{tabular}{lccc}
\toprule
Judge & Overall Acc. & Acc@0 & Acc@1 \\
\midrule
LlamaGuard-3 & 0.54 & 0.17 & 0.34 \\
HarmBench    & 0.59 & 0.30 & 0.45 \\
AegisGuard   & 0.51 & 0.23 & 0.35 \\
JailJudge    & 0.56 & 0.18 & 0.37 \\
\bottomrule
\end{tabular}
\end{sc}
\end{small}
\vspace{0.1cm}
\caption{Leave-one-out evaluation on JST. Accuracy is reported for the held-out judge on samples where exactly zero (\texttt{Acc@0}) or one (\texttt{Acc@1}) of the remaining judges were correct.}\label{tab:judge_stress_test}
\vspace{-0.5cm}
\end{table}

\begin{tcolorbox}[
    colback=white, colframe=customblue, coltitle=white, fonttitle=\bfseries, 
    rounded corners, enhanced, 
    title=Takeaway, 
    attach boxed title to top left={yshift=-2mm, xshift=5mm}, 
    boxed title style={colback=customblue, rounded corners},
    boxsep=1mm,
    left=1mm,
    right=1mm
]
Targeted data selection can improve evaluation reliability
\end{tcolorbox}

%% file: chapter/5_conclusion.tex
\section{Conclusion}

\textbf{Limitations.} Our study does not account for distribution shifts introduced by specific defense mechanisms, such as adversarial training algorithms~\cite{xhonneux2024efficient, casper2024defending}, paraphrasing, or others. For example, the prominent Circuit Breakers~\cite{zou2024improving} defense trains models to produce high-perplexity, non-semantic outputs when confronted with unsafe content. This may further degrade judge performance. Additionally, while we evaluated a diverse set of attacks and models, our findings are limited to the specific combinations tested; unseen distribution shifts in future attacks or models may pose distinct challenges for automated evaluation. For example, future work could examine the impact of strong model-tampering attacks on judge-hacking~\cite{schwinn2024soft}.

In this work, we demonstrate that the current reliance on LLM-as-a-Judge for safety evaluation is flawed. Our human audit reveals that under the distribution shifts inherent to adversarial robustness evaluations, state-of-the-art judges perform, on average, no better than a random coin flip. We show that these inconsistencies can distort attack success rates, suggesting that some reported gains in attack effectiveness, such as for BoN, can be attributed to judge insufficiencies. To help researchers address these limitations in future works, we introduce ReliableBench, a more reliable dataset for safety evaluations, and JudgeStressTest, a challenging benchmark dataset with ground-truth human labels to evaluate the reliability of future judges.

%% file: chapter/appendix.tex





\section{Extended Results}\label{app:extended_results}

Here, we provide additional results and visualizations not included in the main text.

\subsection{Judge Accuracy Heatmaps}
\begin{figure}[h!]
    \centering
    \subfigure[\gemmaoneb{}]{ 
        \includegraphics[width=0.45\linewidth]{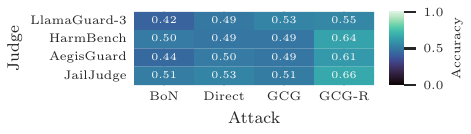}
    }
    \subfigure[\qwen{}]{ 
        \includegraphics[width=0.45\linewidth]{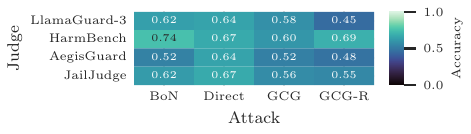}
    }
    \caption{Judge accuracy heatmaps for \gemmaoneb{} and \qwen{}. Darker colors indicate lower accuracy.}
    \label{fig:app_heatmap}
\end{figure}

In Figure~\ref{fig:app_heatmap}, we provide additional visualizations of judge accuracy across different attacks for \gemmaoneb{} and \qwen{}. Consistent with the main text, we observe significant variability in judge performance across different attacks and victim models, further highlighting the impact of attack and model-related shifts.

\subsection{Semantic Categories}\label{app:semantic}

We use the following abbreviations for the semantic harm categories of HarmBench~\citep{mazeika2024harmbench}. CB stands for chemical\_biological, CI represents cybercrime\_intrusion, and HB refers to harassment\_bullying. Additionally, HF is short for harmful, ILG denotes illegal, and MD indicates misinformation\_disinformation.

\subsection{ROC Curves}\label{ROC Curves}
\begin{figure}[h!]
    \centering
    \subfigure[HarmBench]{
        \includegraphics[width=0.23\linewidth]{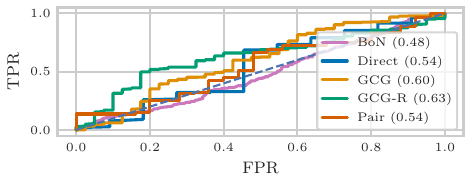}
    }
    \subfigure[JailJudge]{
        \includegraphics[width=0.23\linewidth]{chapter/figures/roc/JailJudge_Llama-3_1.pdf}
    }
    \subfigure[LlamaGuard-3]{
        \includegraphics[width=0.23\linewidth]{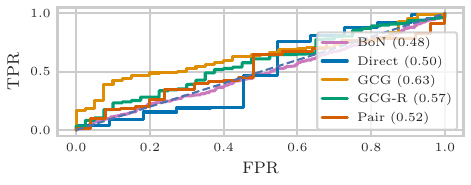}
    }
    \subfigure[AegisGuard]{
        \includegraphics[width=0.23\linewidth]{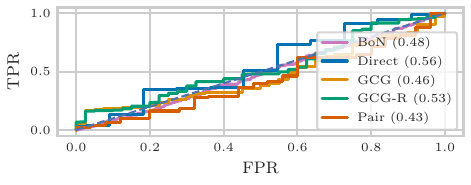}
    }
    
    \subfigure[HarmBench]{
        \includegraphics[width=0.23\linewidth]{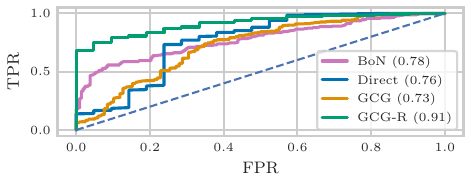}
    }
    \subfigure[JailJudge]{
        \includegraphics[width=0.23\linewidth]{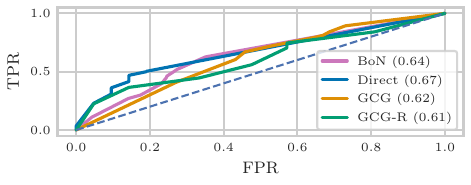}
    }
    \subfigure[LlamaGuard-3]{
        \includegraphics[width=0.23\linewidth]{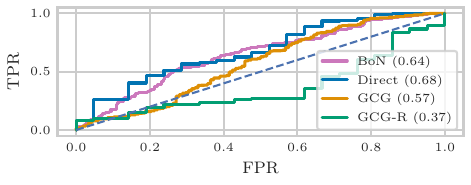}
    }
    \subfigure[AegisGuard]{
        \includegraphics[width=0.23\linewidth]{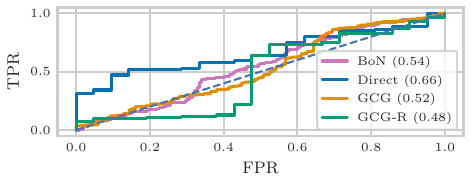}
    }
    \caption{ROC curves for all evaluated judges on \llamaeightb{} and \qwen{}.}
    \label{fig:app_roc_curves}
\end{figure}

We present extended ROC curve analysis for \llamaeightb{} (first row) and \qwen{} (second row) in Figure~\ref{fig:app_roc_curves}. The curves consistently show that across different judges and models, performance often hovers near the diagonal, indicating poor classification performance under distribution shifts present in adversarial safety evaluations. However, larger models such as \qwen{} tend to be easier to judge than smaller models such as \llamaeightb{}.

\subsection{Correlation with Human Ratings}\label{app:correlation}
\begin{figure}[h!]
    \centering
    \subfigure[\gemmaoneb{}]{
        \includegraphics[width=0.22\linewidth]{chapter/figures/correlation_bar/llm_vs_human_bar_Average_Gemma-1B.pdf}
    }
    \subfigure[\gemmalarge{}]{
        \includegraphics[width=0.22\linewidth]{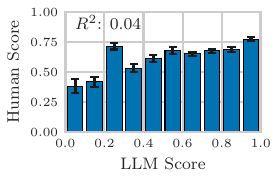}
    }
    \subfigure[\llamaeightb{}]{
        \includegraphics[width=0.22\linewidth]{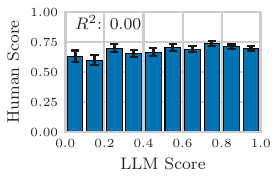}
    }
    \subfigure[\qwen{}]{
        \includegraphics[width=0.22\linewidth]{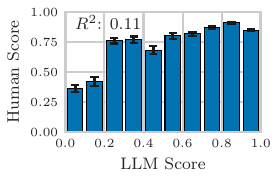}
    }
    \caption{Correlation between human ratings and averaged judge scores for all evaluated models.}
    \label{fig:app_correlation}
\end{figure}

We further analyze the correlation between averaged judge scores and human ratings across all four victim models in Figure~\ref{fig:app_correlation}.

\subsection{Expected True Positives Analysis}\label{app:expected}
We extend our analysis of the probability of finding at least one true positive given $N$ judge-positive samples to all four victim models. The results consistently show that for certain attacks, particularly those involving sampling (e.g., BoN), a single judge-positive indication provides low confidence of actual success, necessitating larger sample sizes for reliable verification.

\begin{figure}[h!]
    \centering
    \subfigure[\gemmaoneb{}]{
        \includegraphics[width=0.22\linewidth]{chapter/figures/expected/expected_harmful_strongreject_gemma-1b.pdf}
    }
    \subfigure[\gemmalarge{}]{
        \includegraphics[width=0.22\linewidth]{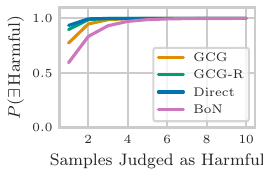}
    }
    \subfigure[\llamaeightb{}]{
        \includegraphics[width=0.22\linewidth]{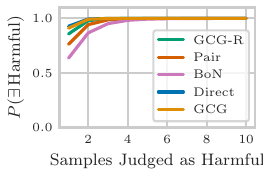}
    }
    \subfigure[\qwen{}]{
        \includegraphics[width=0.22\linewidth]{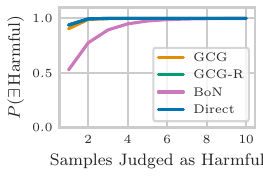}
    }
    \caption{Probability of finding at least one true positive given $N$ judge-positive samples across all models.}
    \label{fig:app_expected_harmful}
\end{figure}

\subsection{Attack Optimization}\label{app:attackopt}

Our results in Section~\ref{sec:attack_opt} support a recent hypothesis from~\citet{beyer2025sampling} that adversarial attacks focus on breaking the model's refusal mechanism rather than making the resulting content more harmful. By filtering for samples labeled as \enquote{positive} by the StrongReject judge, we likely increased the number of non-refusals in our dataset, which explains why harmfulness is already relatively high at the start of Figure~\ref{fig:correlation_steps} and does not increase as optimization continues. Our results provide additional evidence that current attack optimizers do not, in fact, cause models to generate more harmful content during optimization.

\subsection{Concordance Scores}\label{app:concordance}

\begin{figure}[h]
    \centering
    \includegraphics[width=0.5\linewidth]{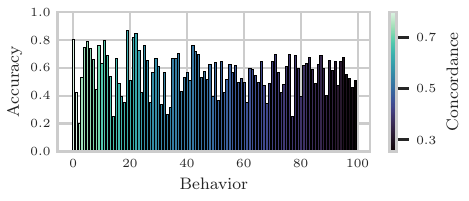}
    \caption{Average accuracy vs. inter-judge agreement across 100 behaviors. The height of each bar represents the judges’ average accuracy for a given behavior category. The color encodes the mean concordance score for that behavior.}\label{fig:app_judge_concordance_score_heatmap}
\end{figure}

To investigate if consensus implies correctness, we evaluate the relationship between inter-judge agreement (concordance) and objective accuracy across 100 behaviors. Figure~\ref{fig:app_judge_concordance_score_heatmap} displays these metrics, categorized by behavior.

The data reveal that high concordance (light color) does not consistently correlate with high accuracy (bar height). This lack of a clear trend confirms that judge agreement is an insufficient proxy for ground-truth correctness when using existing judges.

\subsection{ReliableBench}\label{app:reliablebench}

\begin{figure}
    \centering
\subfigure[Accuracy Distribution]{
        \includegraphics[width=0.5\linewidth]{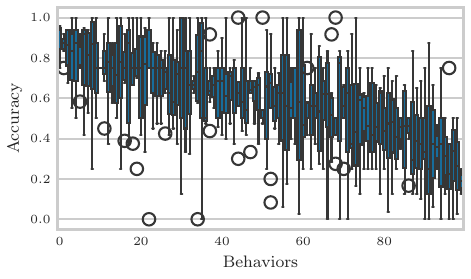}
    }
    \caption{Analysis of judge reliability for the construction of ReliableBench. Boxplots show the distribution of accuracy across different attacks and models, confirming that the selected subset remains robustly easier to judge despite variations in the evaluation setting.}
    \label{fig:reliable_bench_appendix}
\end{figure}

To verify the robustness of these findings, we analyze judge accuracy across all evaluated model-attack pairs (Figure~\ref{fig:reliable_bench_appendix}). Despite variability across specific configurations, the consistent decrease in mean accuracy as more behaviors are included confirms that the identified subset ReliableBench is robustly easier to judge.

